\documentclass[sigconf]{acmart}
\AtBeginDocument{%
  }

\newcommand*{\ie}{i.e.,\xspace}
\newcommand*{\eg}{e.g.,\xspace}
\copyrightyear{2025}
\acmYear{2025}
\acmConference[FAccT '25]{The 2025 ACM Conference on Fairness, Accountability, and Transparency}{June 23--26, 2025}{Athens, Greece}
\acmBooktitle{The 2025 ACM Conference on Fairness, Accountability, and Transparency (FAccT '25), June 23--26, 2025, Athens, Greece}\acmDOI{10.1145/3715275.3732105}
\acmISBN{979-8-4007-1482-5/2025/06}

\usepackage{graphicx}
\usepackage{subcaption}
\usepackage{multirow}
\usepackage{soul}
\usepackage{xspace}
\usepackage{xcolor}

\renewcommand*{\eg}{e.g.,\xspace}

\begin{document}

\title{SHAP-based Explanations are Sensitive to Feature Representation}


\author{Hyunseung Hwang}
\affiliation{
\institution{KAIST}
\city{Daejeon}
\country{Republic of Korea}
}
\email{aguno@kaist.ac.kr}

\author{Andrew Bell}
\affiliation{
\institution{New York University}
\city{New York}
\state{NY}
\country{USA}
}
\email{alb9742@nyu.edu}

\author{Joao Fonseca}
\affiliation{
\institution{New York University}
\city{New York}
\state{NY}
\country{USA}
}
\email{jpm9748@nyu.edu}

\author{Venetia Pliatsika}
\affiliation{
\institution{New York University}
\city{New York}
\state{NY}
\country{USA}
}
\email{venetia@nyu.edu}

\author{Julia Stoyanovich}
\affiliation{
\institution{New York University}
\city{New York}
\state{NY}
\country{USA}
}
\email{stoyanovich@nyu.edu}

\author{Steven Euijong Whang}
\affiliation{
\institution{KAIST}
\city{Daejeon}
\country{Republic of Korea}
}
\email{swhang@kaist.ac.kr}

\renewcommand{\shortauthors}{Hwang et al.}

\begin{abstract}

Local feature-based explanations are a key component of the XAI toolkit. These explanations compute feature importance values relative to an ``interpretable'' feature representation. In tabular data, feature values themselves are often considered interpretable. This paper examines the impact of data engineering choices on local feature-based explanations. We demonstrate that simple, common data engineering techniques, such as representing age with a histogram or encoding race in a specific way, can manipulate feature importance as determined by popular methods like SHAP. Notably, the sensitivity of explanations to feature representation can be exploited by adversaries to obscure issues like discrimination. While the intuition behind these results is straightforward, their systematic exploration has been lacking. Previous work has focused on adversarial attacks on feature-based explainers by biasing data or manipulating models. To the best of our knowledge, this is the first study demonstrating that explainers can be misled by standard, seemingly innocuous data engineering techniques.

\end{abstract}




\maketitle

\section{Introduction}
\label{sec:introduction}

Explainable AI (XAI) is becoming increasingly critical for justifying the behavior of AI systems implemented in high stakes domains like education, lending, public employment, and healthcare~\cite{zejnilovic2021machine, alwarthan2022explainable, chaddad2023survey}. One of the key components of XAI is the use of \emph{local feature-based explanations}, which quantify the importance of an observation's features to an outcome (or some other quantity of interest).  For example, these explanations are foundational for \emph{algorithmic recourse}, where understanding \emph{why} an individual is rejected for a loan by an AI-assisted system allows them to contest or reverse that unfavorable decision. Local feature-based explanations can also be used to surface unfairness in decision-making, for example, if a model is revealed to be making individual-level decisions on the basis of features like age, gender, or race, which may be illegal to use under the disparate treatment doctrine.\footnote{\url{https://en.wikipedia.org/wiki/Disparate_treatment}}  Further, these explanations are becoming an essential tool to fulfill legal and regulatory requirements, such as the European Union's AI Act, and General Data Protection Regulation's ``right to explanation''~\cite{fresz2024should}.

The Shapley value framework~\cite{shapley1953}, originally developed for dividing revenue in cooperative games, is widely used to quantify local feature importance in predictive classification. It underpins prominent explanation methods like SHAP (Shapley Additive Explanations)~\cite{DBLP:conf/nips/LundbergL17} and QII (Quantitative Input Influence)~\cite{DBLP:conf/sp/DattaSZ16}. The framework explains the classification outcome for an observation by assessing how changes to a feature's value, individually or in combination with others, impact that outcome. This process simulates interventions, aligning with causal inference principles by isolating each feature's influence while controlling for others. A high Shapley value for a protected feature like age suggests its significant influence on the classifier's decision.

However, Shapley-value-based explanations have limitations: they can mislead users (intentionally or unintentionally)~\cite{huang2024failings} and are vulnerable to adversarial attacks and manipulations~\cite{DBLP:conf/iclr/LabergeA0MK23}. \textbf{In this paper, we focus on the key observation that local feature-based explanations, derived from trained models and post-processed data, are susceptible to manipulations through feature engineering, which occurs upstream from classification in the machine learning pipeline.} We use SHAP~\cite{DBLP:conf/nips/LundbergL17}, the most widely adopted implementation of the Shapley value framework, to show that local feature-based explanations are influenced by simple data engineering operations, such as transforming continuous values or encoding categorical values, which modify feature representations.  For instance, bucketization---a common method of grouping values into ranges---can make a feature value appear less (or more) important in terms of SHAP.  We now provide an intuition through the example below:

\begin{figure}[t]
\centering
    \includegraphics[width=\columnwidth]{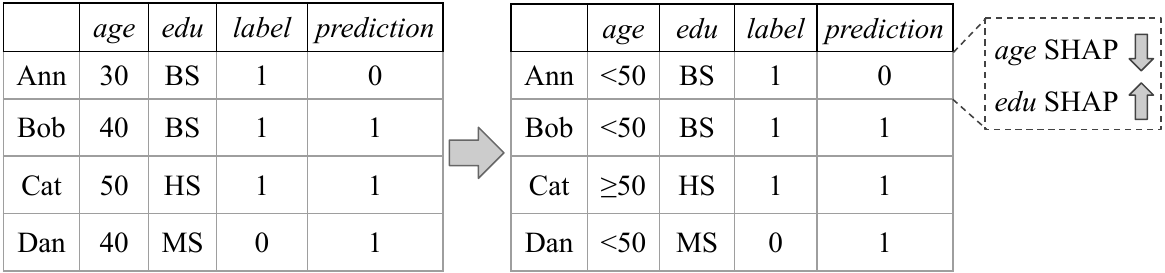}
\caption{A hypothetical lending example: For a given classifier model, if we bucketize the \emph{age} feature to generate a local explanation of the outcome with SHAP, then the importance of \emph{age} for Ann decreases compared to using the raw value of the feature.  Intuitively, this happens because $age=30$ is infrequent---and low---in this hypothetical dataset, while $age < 50$ appears to be typical, both on its own and in combination with education.  
}
\label{fig:motivation}
\vspace{-0.27cm}
\end{figure}

\begin{figure}[t]
    \fbox{
    \begin{subfigure}[t]{0.45\textwidth}
        
        \includegraphics[width=1.0\linewidth]{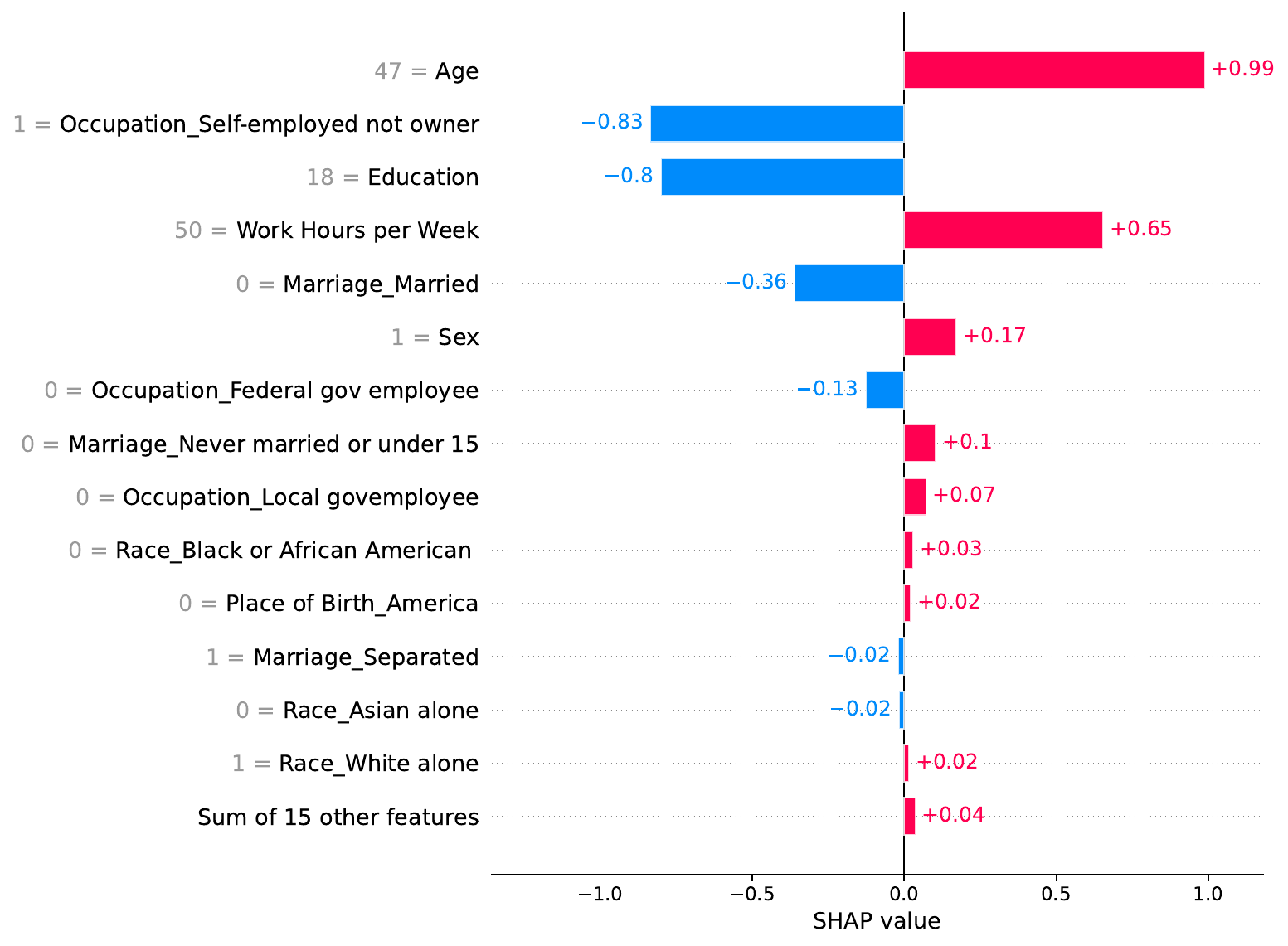}
        \caption{SHAP values before bucketization of \texttt{age}}
        \label{fig:individualanalysis:a}
    \end{subfigure}
    }
    \fbox{
    \begin{subfigure}[t]{0.45\textwidth}
        \includegraphics[width=1.0\linewidth]{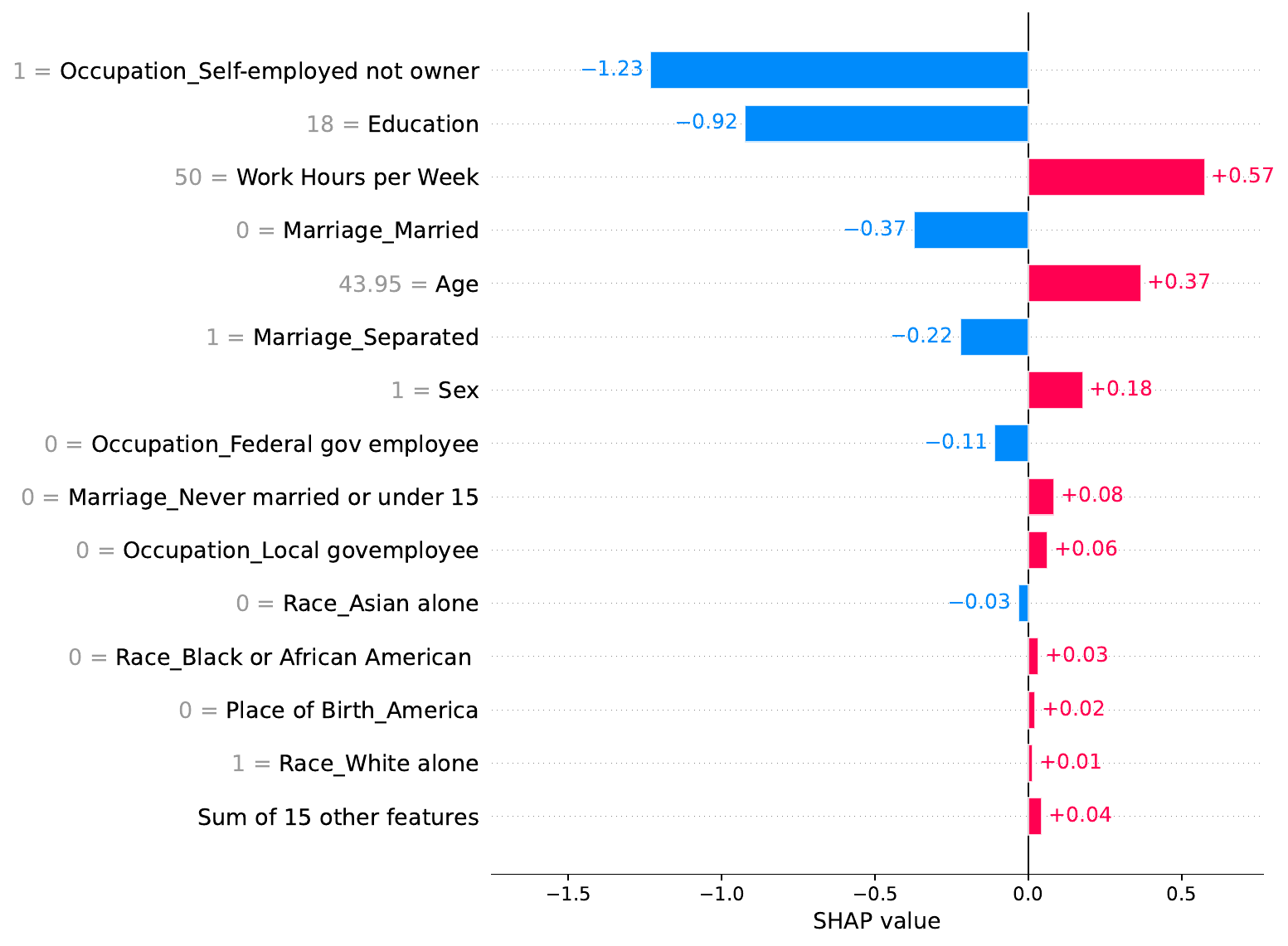}
        \caption{SHAP values after \texttt{age} is bucketized}
        \label{fig:individualanalysis:b}
    \end{subfigure}
    }
    \caption{SHAP values of features before (a) and after (b) bucketization for a fixed individual from the ACS Income dataset. Note that the classifier model and SHAP explainer remain fixed; the only modification to the individual's features from (a) $\rightarrow$ (b) was the bucketization of age. In (a), age is encoded as a continuous feature and is deemed most important by SHAP, with a rank of 1 and a feature weight of 0.99. In (b), the age feature was bucketized into 12 equi-width intervals over its active domain, using the median age to represent observations within each interval. This decreased the feature weight to 0.37, demoting age to the 5th rank in importance.}
    \label{fig:individualanalysis}
\end{figure}

\begin{example}[Motivating example]
Consider a vendor --- a financial institution that uses a binary classifier to approve loans (see Figure~\ref{fig:motivation}). Suppose that Ann applies for the loan and is incorrectly rejected (a false negative).  The vendor would like to see if its model made this rejection decision based on Ann's age, and decides to compute feature importance using SHAP as part of its analysis.  Indeed, when SHAP is run over the raw feature values, age appears to have high importance, likely because an age of 30 is comparatively low in the vendor's data.  Worried about a potential lawsuit, the vendor attempts to generate a different explanation for the same classification outcome: they keep the classifier model fixed, but change the representation of age, ``bucketizing'' it into the ranges ``below 50'' and ``50 and above''.  The vendor is relieved to see that this simple manipulation substantially diminishes the importance of age when explaining Ann's outcome. 

Figure~\ref{fig:individualanalysis} demonstrates this very scenario for an individual in the ACS Income dataset, with the SHAP plot on the top showing an explanation on raw feature values, and the plot on the bottom showing an explanation after age is bucketized.   Observe that the importance of age drops from rank 1 (most important) in Figure~\ref{fig:individualanalysis:a} to rank 5 (somewhat important) in Figure~\ref{fig:individualanalysis:b}, a decrease of 5 positions in terms of importance.
Note also that, because of the efficiency property of Shapley values~\cite{shapley1953}, a SHAP explanation can be used to reconstruct the outcome (by summing feature weights and returning the positive label if the sum is positive). In the example in Figure~\ref{fig:individualanalysis}, both explanations are consistent with the classifier's prediction: they both predict that the individual would be rejected for the loan. If the vendor is worried about being challenged for using a protected feature like age to incorrectly reject applicants, it can look for an explanation that agrees with the prediction, but diminishes the importance of age. \textbf{In this paper, we refer to this kind of a manipulation as a data engineering attack.}
\end{example}

        
\paragraph{Contributions and roadmap}

In this paper, we systematically investigate how SHAP-generated feature-based explanations are affected by simple data engineering choices, and how this sensitivity can be used to design a data engineering attack on SHAP. 
We discuss related work in Section~\ref{sec:relatedwork}, describe preliminaries in Section~\ref{sec:preliminaries}, and then present our contributions.

\begin{itemize}
\item As our first contribution, in Section~\ref{sec:sensitivity}, we empirically examine the impact of bucketization or binning on continuous features (\eg age) and of different encoding methods on categorical features (\eg race). We show that SHAP is highly sensitive to data engineering choices, with the importance of age changing by as much as 20 rank positions in some cases.  Further, in cases where age is the most important feature, its importance frequently drops by between 3 and 5 positions in the ranking. When using the race feature, we show that merging White and Asian individuals or White and Black individuals into a single category can reduce the importance of the race feature to nearly 0.

\item As our second contribution, in Section~\ref{sec:attack}, we design a feature engineering attack, demonstrating that sensitivity to seemingly benign data engineering choices can enable adversarial vendors to obscure the importance of protected features with minimal impact on predictions, allowing them to evade scrutiny without model retraining the model.  For example, we demonstrate that our attack generally outperforms equi-width bucketization by substantially reducing the importance of the age feature without sacrificing explanation fidelity.


\end{itemize}

In Section~\ref{sec:discussion}, we highlight the need for a more robust framework for model explanations that evaluates not only accuracy and fairness, but also the impact of data engineering on local feature-based explanations. \textbf{Creating tools and guidelines to ensure that data engineering choices do not unduly influence reported feature importance should become standard practice in AI development.} We also acknowledge limitations and outline future directions. Finally, in Section~\ref{sec:conclusion}, we summarize our insights.

All code is available at \url{https://github.com/Aguno/Shap-Attack}.

\section{Related Work}
\label{sec:relatedwork}

The relationship between feature engineering and model explainability has been explored in previous works. For example, \citet{DBLP:conf/aaai/Ribeiro0G18} investigate how feature \emph{selection} and engineering techniques impact model explanations, focusing on how feature importance is derived from global model behavior. Our research complements this approach by systematically demonstrating how bucketization and binning can impact model explanations.

Other studies have examined how data engineering operations like re-scaling, re-weighting, and re-sampling of features can either mitigate or exacerbate bias~\cite{DBLP:journals/kais/KamiranC11,DBLP:journals/jmlr/ZafarVGG19}. These works demonstrate the unintended consequences of seemingly innocuous feature engineering decisions on AI systems. However, none of these studies explicitly address explainability, particularly in the context of SHAP, one of the most widely adopted explanation methods~\cite{belle2021principles, bhatt2020explainable}.

More recently, \citet{DBLP:conf/aies/SlackHJSL20} proposed an adversarial attack on SHAP by scaffolding a classifier that may be unfair on the input data, but appears fair on the rest of the data in terms of common statistical fairness criteria. In another recent approach, \citet{DBLP:conf/aaai/BanieckiB22} generate synthetic data to manipulate SHAP. The authors use a genetic algorithm to manipulate the feature values towards certain SHAP targets. Finally, the Fool SHAP method by \citet{DBLP:conf/iclr/LabergeA0MK23} uses biased sampling to construct the background data such that the protected feature's importance is reduced, allowing the vendor to show false compliance  during an algorithmic fairness audit.  Here, an optimization problem is formulated to reduce the SHAP value of a feature without significantly altering the background data distribution relative to the original data.  However, sampling data can be viewed as an explicit manipulation and may be prohibited. In comparison, our work does not focus on model fairness and only performs common data engineering manipulations that a data analyst could legitimately perform.  We show that such seemingly benign operations can alter SHAP-computed feature importance and be used to design an attack.

Unlike prior work on predictive multiplicity~\cite{DBLP:conf/fat/BlackRB22}, starting with the work on the ``Rashomon Effect''~\cite{breiman2001statistical}, where different models may have comparable performance, our contribution is in showing that the same model may produce different explanations due to seemingly innocuous data preprocessing (i.e., feature engineering) choices. This exposes a critical weakness in SHAP that has not been shown in prior work and that, as we demonstrate, can enable an adversarial actor to manipulate explanations.

Finally, there are other works that interrogate explainability ~\cite{DBLP:conf/fat/BarocasSR20,DBLP:conf/fat/Hancox-LiK21,DBLP:conf/fat/Hancox-Li20} (e.g., counterfactual explanations), but they do not identify data preprocessing vulnerabilities when using SHAP as we do.
\section{Preliminaries}
\label{sec:preliminaries}

\subsection{Local feature-based explanations with Shapley values}

The Shapley value framework~\cite{shapley1953} is widely used to quantify local feature importance in predictive models~\cite{DBLP:conf/sp/DattaSZ16,DBLP:conf/nips/LundbergL17}. It does so by attributing a model’s output for a given instance to individual input features, based on how their inclusion---alone or in combination with other features---affects the prediction.

The Shapley value for a feature \( i \) is formally defined as:
\[
\phi_i(f) = \sum_{S \subseteq N \setminus \{i\}} \frac{|S|!(|N|-|S|-1)!}{|N|!} \left( f(S \cup \{i\}) - f(S) \right)
\]
where \( N \) is the set of all players (features), \( S \subseteq N \setminus \{i\} \) is a subset of players excluding player \( i \), \( f \) is a value function defined on subsets of \( N \) (e.g., the expected model output conditional on the features in \( S \)), \( \phi_i(f) \) is the Shapley value assigned to player \( i \), representing their marginal contribution averaged over all possible coalitions.

In predictive classification, the players correspond to input features, and \( f(S) \) is typically defined as the expected value of the model output conditional on the feature values in subset \( S \). The vector of Shapley values assigned to an instance’s features constitutes an \emph{explanation}. Due to the efficiency property of Shapley values~\cite{shapley1953}, the sum of these contributions exactly recovers the model output (minus a baseline), ensuring additive consistency.

By convention, a feature's importance is indicated by the absolute value of its weight (with higher values denoting greater importance), while the sign reflects the direction of its contribution toward a specific prediction (positive or negative). For example, in Figure~\ref{fig:individualanalysis}(a), the \textit{age} feature has a weight of $-0.59$: its high absolute value indicates importance, and the negative sign points toward the negative class label. Also by convention, visual explanations sort features in descending order of absolute weight, with the most important feature ranked first, regardless of sign, followed by the next most important, and so on.

\begin{figure*}[t]
    \begin{subfigure}[t]{0.28\textwidth}
        \includegraphics[width=1.0\linewidth]{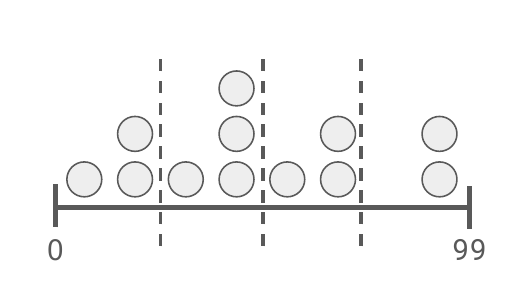}
        \caption{4 equi-width buckets}
        \label{fig:bucket_explanations:a}
    \end{subfigure}
    \begin{subfigure}[t]{0.28\textwidth}
        \includegraphics[width=1.0\linewidth]{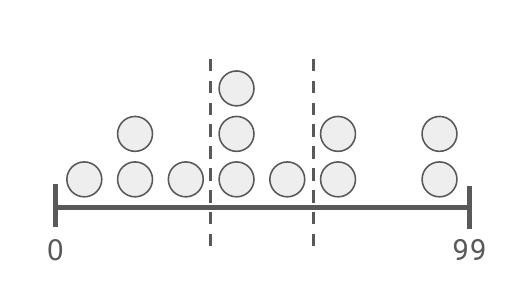}
        \caption{3 equi-depth buckets}
        \label{fig:bucket_explanations:b}
    \end{subfigure}
    \begin{subfigure}[t]{0.28\textwidth}
        \includegraphics[width=1.0\linewidth]{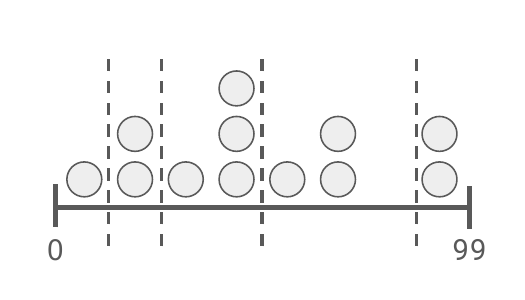}
        \caption{custom buckets per Section~\ref{sec:manipulation}}
        \label{fig:bucket_explanations:c}
    \end{subfigure}
    \caption{The figures above represent different ways to bucketize a continuous or high-dimensional ordinal feature like \textit{age}. In each sub-figure, the feature is represented as a lineplot with values from 0 to 99. Each circle represents an age value for a single observation from the dataset, and the observations are the same across the sub-figures. For equi-width buckets (a), the domain is divided into buckets of equal width. For equi-depth buckets (b), buckets are created that all contain an approximately equal number of observations (this is equivalent to equi-width buckets over the percentile values of the feature). Sub-figure (c) shows how custom buckets may be created by the method described in Section~\ref{sec:manipulation} to manipulate the SHAP rank of a feature.}
    \label{fig:bucket_explanations}
\end{figure*}

In this work, we use SHAP~\cite{DBLP:conf/nips/LundbergL17}, with its open-source implementation\footnote{\url{https://pypi.org/project/shap/}}. SHAP is used extensively by industry practitioners~\cite{DBLP:journals/ese/UmmeHabibaHBFW25,DBLP:journals/ais/IbanezO22}, highlighting a critical need for its continued study, particularly when it comes to surfacing vulnerabilities that could be abused by bad actors. We reveal the sensitivity of
SHAP to feature engineering, adding to a robust body of work on studying the tool, surfacing issues, patching them, or expanding SHAP in other ways towards better implementation in practice~\cite{DBLP:conf/fat/WexlerPRBZ20,DBLP:conf/fat/Vengroff24,DBLP:conf/fat/Watson22}. 

Many researchers and practitioners use XAI for fairness auditing.  Most relevant to our work, a high Shapley value for a protected feature like \emph{age} or \emph{race} suggests the feature has a significant influence on the classification outcome, which may raise ethical or legal concerns.  \citet{DBLP:conf/fat/WexlerPRBZ20} present an approach for using SHAP and ``what-if tools'' to probe ML models for fairness. \citet{DBLP:conf/fat/Vengroff24} develops a toolkit based on SHAP that helps identify bias in both an ML system and the data used to train that system. \citet{DBLP:conf/fat/DeckSD024} offer a nuanced perspective, noting that XAI tools are not an ``ethical panacea,'' but are ``one of many tools to approach the multidimensional, sociotechnical challenge of algorithmic fairness,'' along with other tools like those focused on bias auditing.

\subsection{Representing features}
\label{sec:representationimpact}

In tabular data, features can be continuous, ordinal, or categorical. Continuous data can take any value within a range and is measurable, while ordinal data represents categories with a meaningful order or ranking. 
Categorical data consists of distinct groups or categories without an inherent ordering. 

There are various ways to represent these data types in machine learning pipelines. For continuous or ordinal features, we may leave the data as is, or discretize the values using bucketization, where values are grouped into ranges. These buckets can then be one-hot encoded or treated as ordinal features. Categorical features can be encoded using methods like one-hot, ordinal, or target encoding. Additionally, scaling or normalization may be applied to adjust distributions or ranges of features.

\paragraph{Encoding continuous or ordinal features.} Consider the feature \textit{age}, which may be continuous or ordinal with integer values. For prediction or explanation purposes, \textit{age} could be used in its raw form, bucketized into ordinal categories, or encoded into multiple buckets via one-hot encoding. Importantly, ``upstream'' feature representation choices affect the properties of the ``downstream'' model in a machine learning pipeline~\cite{DBLP:journals/cacm/StoyanovichAHJS22}, influencing its accuracy~\cite{jokanovic2016effect}, fairness, and explainability.

In this work, we focus on how bucketization~\cite{DBLP:conf/vldb/Ioannidis03}---grouping continuous or ordinal data into distinct value ranges---affects model predictions and SHAP explanations. We explore three methods for bucketizing continuous features, as illustrated in Figure~\ref{fig:bucket_explanations}. The first, \emph{equi-width} in Figure~\ref{fig:bucket_explanations:a}, creates buckets of equal feature value ranges. The second, \emph{equi-depth} in Figure~\ref{fig:bucket_explanations:b}, ensures each bucket contains an approximately equal number of data points (\ie the buckets represent percentiles of the data). The third method, in Figure~\ref{fig:bucket_explanations:c}, employs Bayesian Optimization to define bucket widths that optimize an adversarial objective, which we describe in Section~\ref{sec:manipulation}.  Note that the number of buckets can vary across methods.


\paragraph{Categorical features.} We frame our discussion of encoding categorical features through the protected feature \textit{race}. Whenever practitioners include \textit{race} as a feature in machine learning models, they are implicitly making choices about how to encode that feature~\cite{benthall2019racial, wang2022towards}. For example, the ACS Income and ACS Public Coverage datasets\footnote{\url{https://github.com/socialfoundations/folktables}}, used in our experiments, include eight distinct race categories plus a null value. One approach is to represent all eight categories using one-hot encoding. However, due to small sample sizes in some categories, practitioners often create an ``other'' supercategory, leading to arbitrary groupings. Further, one category represents individuals identifying as ``mixed race,'' but lacks information on which races they identify with. Intersectional encodings could also be considered.

In this work, we focus on six plausible encoding methods of the \textit{race} feature, shown in Table~\ref{tbl:buckets}.  Four of these individuals into two race categories and two split them into three categories. While not exhaustive, these encodings are sufficient for exploring the sensitivity of SHAP to different race encodings. 

\textbf{Note:} One-hot encoding can lead to counter-intuitive or redundant explanations.  Consider, for example, a simple binary feature that denotes whether a person is a smoker.  This feature would be represented by two one-hot-encoded features: \emph{smoker=yes} (set to 0 for a non-smoker) and \emph{smoker=no} (set to 1 for a non-smoker).  An explanation of a medical diagnosis may redundantly assign high importance to both \emph{smoker=yes} and \emph{smoker=no}: a person may be predicted to have a low likelihood of developing lung cancer both because they are a non-smoker (\emph{smoker=no} is set to 1) and because they are not a smoker (\emph{smoker=yes} is set to 0).  An explanation may also be counter-intuitive, assigning high importance to \emph{smoker=yes} being set to 0 and low importance to \emph{smoker=no} being set to 1.

Returning to the one-hot representation of \emph{race}: an explanation of a racially biased lending decision may assign a high positive weight to both \emph{race=White} being set to 1 (the applicant is White), and to \emph{race=Black} being set to 0 (the applicant is not Black).  To estimate the total impact of race on the outcome, we sum up the weights of all one-hot-encoded components of the feature.

\begin{figure*}[t]
    \begin{subfigure}[t]{0.32\textwidth}
        \includegraphics[width=\linewidth]{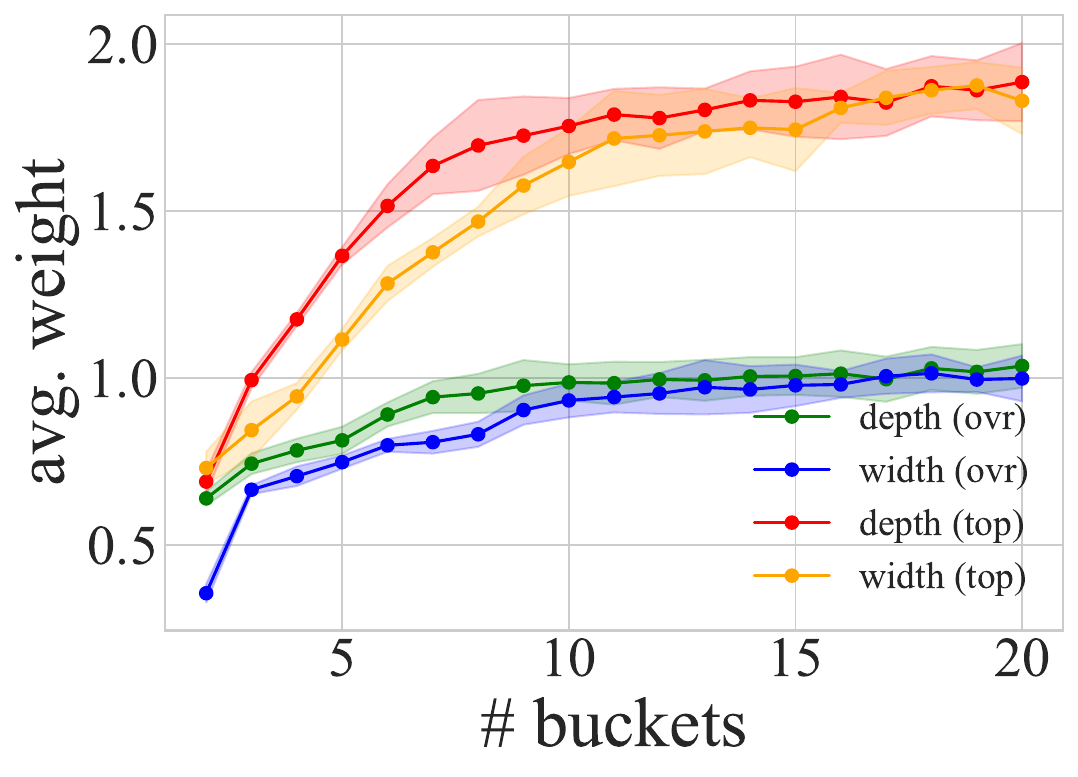}
        \caption{Average SHAP value for the overall population (ovr) and subset of samples with \emph{age} as the most important feature (top).}
        \label{fig:avg-shap-age}
    \end{subfigure}\hfill
    \begin{subfigure}[t]{0.32\textwidth}
        \includegraphics[width=\linewidth]{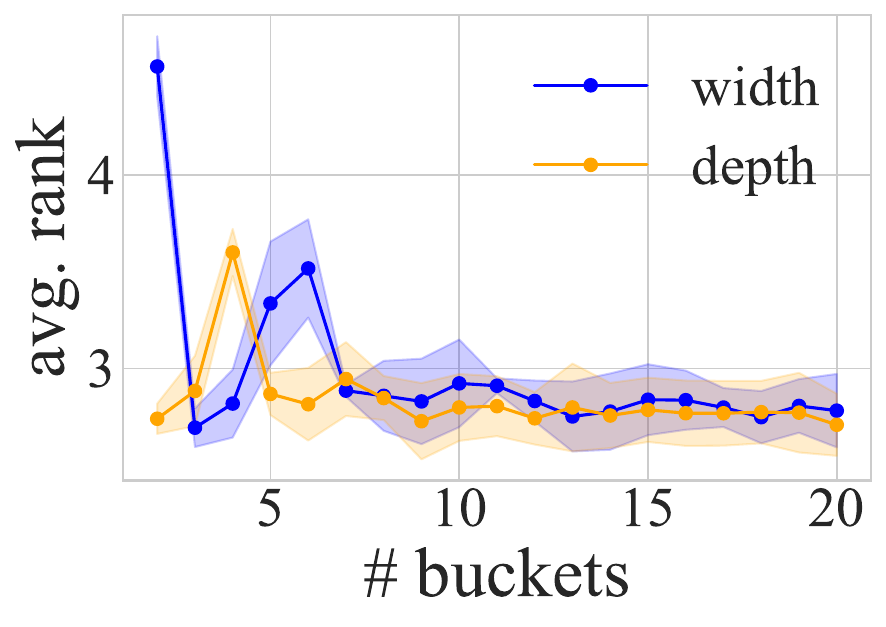}
        \caption{Average rank of \emph{age} in terms of feature importance.}
        \label{fig:avg-rank-age}
    \end{subfigure}\hfill
    \begin{subfigure}[t]{0.32\textwidth}
        \includegraphics[width=\linewidth]{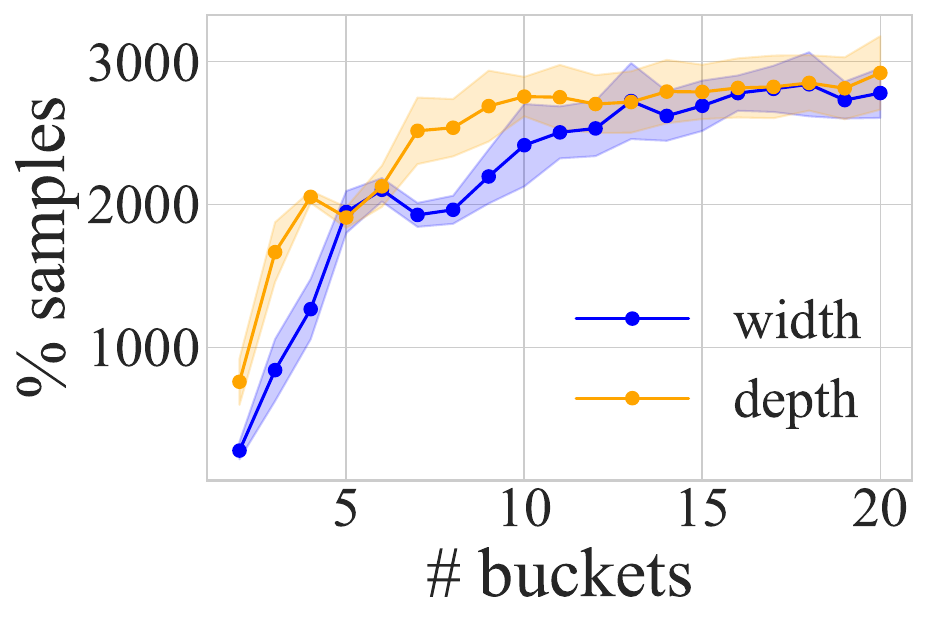}
        \caption{Percentage of samples for which \emph{age} is the most important feature.}
        \label{fig:top1freq-age}
    \end{subfigure}

    \caption{ 
    Number of buckets versus average feature importance weight and rank of the \emph{age} feature on ACS Income. For each plot, we compare uniform (equi-width) and quantile (equi-depth) histograms. 
    }
   
    \label{fig:sensitivity}
\end{figure*}

\subsection{Experimental setup}
\label{sec:preliminaries:exp}

We ran experiments over two real-world benchmark datasets with associated predictive tasks.    

\begin{itemize}
    \item[(1)] ACS Income (Virginia, 2018)\,\cite{DBLP:conf/nips/DingHMS21} is used to predict whether an individual's income is above \$50K. It contains 46,144 observations comprised of 8 features, out of which 5 are categorical. 
    \item[(2)] ACS Public Coverage (Virginia, 2018)\,\cite{DBLP:conf/nips/DingHMS21} is used to predict whether an individual is covered by public health insurance. It contains 25,524 observations comprised of 16 features, out of which 13 are categorical. 
\end{itemize}

For both tasks, we treat \emph{age} and \emph{race} as protected features, and one-hot encode all categorical features, including \emph{race}.   We use XGBoost, a state-of-the-art ensemble classifier, with hyperparemeter tuning for overall accuracy. 


\paragraph{Evaluating explanations.}

SHAP-based explanations can be evaluated using many different metrics~\cite{DBLP:series/hci/Robnik-SikonjaB18,fairxai}. In this work, we use three metrics.  
An explanation is considered faithful if, for a given observation, the sum of its feature importance weights (before or after any preprocessing modifications) corresponds to the originally predicted outcome. The first metric, {\em fidelity}, refers to the proportion of observations for which a faithful explanation was generated, expressed as a ratio of those observations to the total number.

Additionally, we quantify how data representation choices affect both the absolute and the relative importance of a feature. We compute the \emph{average SHAP value} (feature weight) and the \emph{average rank} (by absolute value of feature weight) of the protected feature, and quantify the change in these metrics to compare explanations.  Difference in average SHAP value quantifies the absolute change in feature importance, while difference in average rank quantifies the relative change in feature importance.

\paragraph{Sensitivity versus attack experiments}
In the sensitivity experiments (Section~\ref{sec:sensitivity}), we alter the feature representation in both the training data (used to train the classifier) and the test data (used by the explainer). Here, we examine how explanations respond to feature bucketization, training a new model each time and applying the same representation to both the classifier and explainer.

In contrast, in the attack experiments (Section~\ref{sec:attack}), we train the model on the original, non-bucketized data.  We then keep the model fixed and only modify the representation of the inputs to the explainer. Here, we show that different explanations can be produced for the same model. In particular, this allows us to shift the apparent importance of a sensitive feature without retraining. 

In both settings, the baseline applies no additional feature engineering beyond simple one-hot encoding for categorical features.
\section{First Contribution: Sensitivity of SHAP to Feature Engineering}
\label{sec:sensitivity}

The experimental results reported in this section demonstrate that seemingly trivial feature engineering choices can lead to potentially harmful outcomes. 

\subsection{Continuous features (\emph{age})}
\label{sec:continuousfeaturesage}

We evaluate SHAP's sensitivity to bucketization of the continuous \emph{age} feature on the ACS Income dataset.  Figure~\ref{fig:avg-shap-age} shows that the feature importance of  \emph{age} increases with increasing number of buckets, both overall and for the observations for which \emph{age} is the most important feature when the classifier is trained on raw (unbucketized) data. Figure~\ref{fig:avg-rank-age} complements this result by showing that the average rank of \emph{age}  decreases (\ie \emph{age}  moves closer to the top of the list) with increasing number of buckets.  Figure~\ref{fig:top1freq-age} shows that the percentage of observations for which \emph{age} is the most important feature increases substantially with increasing number of buckets, showing high sensitivity.

Intuitively, as the number of buckets increases, \emph{age} becomes less obfuscated and thus plays a more important role in a model's prediction. Hence, even if bucketization itself is a standard operation, the importance of \emph{age} , both in absolute terms (its weight) and in relative terms (its rank) can change drastically for a substantial portion of the observations. 
Similar to the sensitivity testing scenario, more buckets means the \emph{age} is more fine-grained and has higher SHAP values across the entire dataset. However, the average SHAP value among the observations for which  \emph{age} is the most important feature remains similar regardless of the number of buckets. 

\begin{figure*}[t]
   \begin{subfigure}[t]{0.35\textwidth}
       \includegraphics[width=\linewidth]{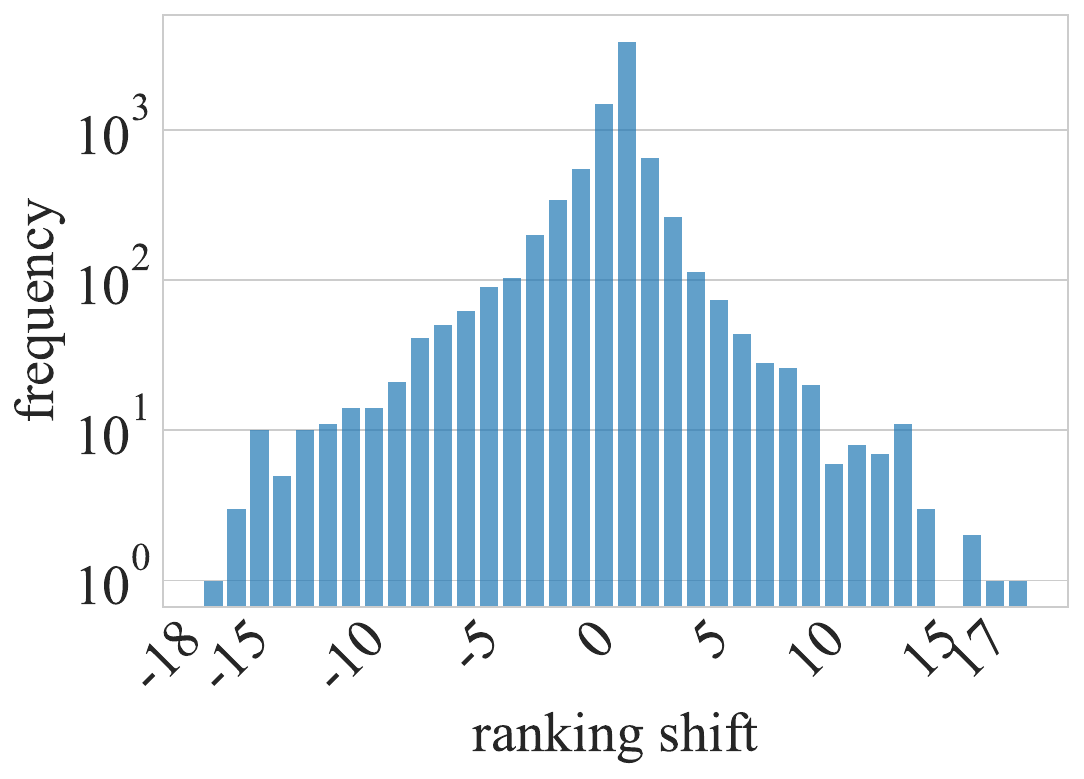}
       \caption{5 equi-width buckets rank shifts}
   \end{subfigure}
   \begin{subfigure}[t]{0.35\textwidth}
       \includegraphics[width=\linewidth]{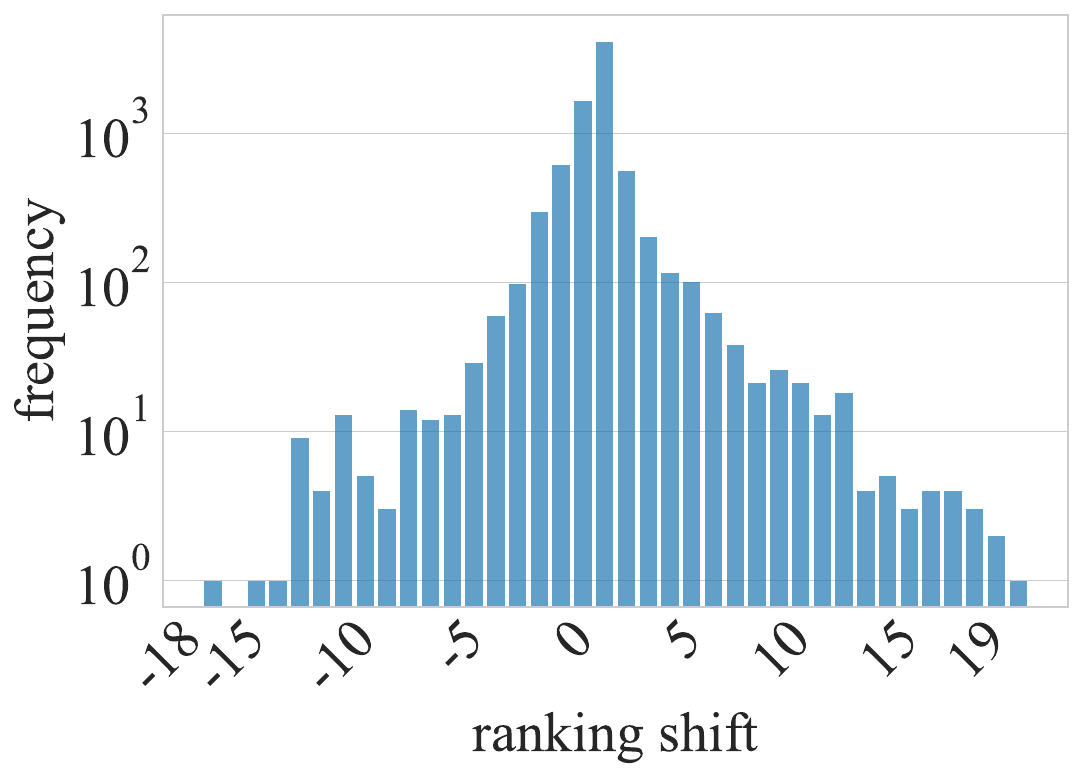}
       \caption{5 equi-depth buckets rank shifts}
   \end{subfigure} \\
   \begin{subfigure}[t]{0.35\textwidth}
       \includegraphics[width=\linewidth]{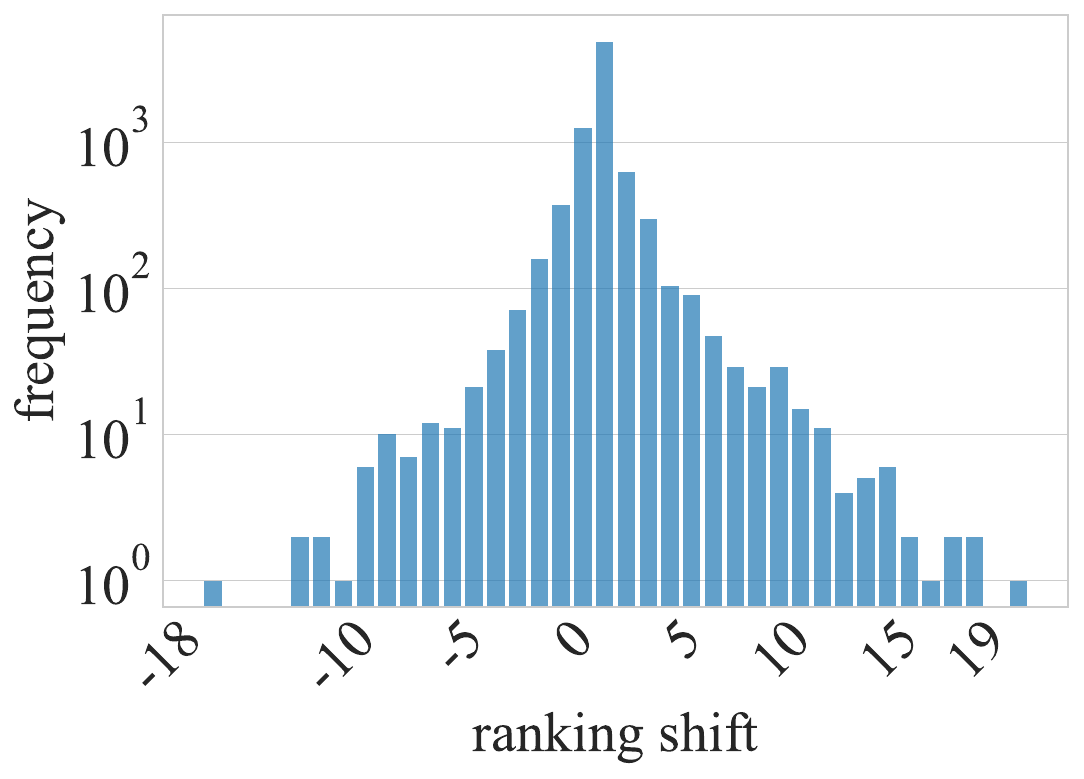}
       \caption{10 equi-width buckets rank shifts}
   \end{subfigure}
   \begin{subfigure}[t]{0.35\textwidth}
       \includegraphics[width=\linewidth]{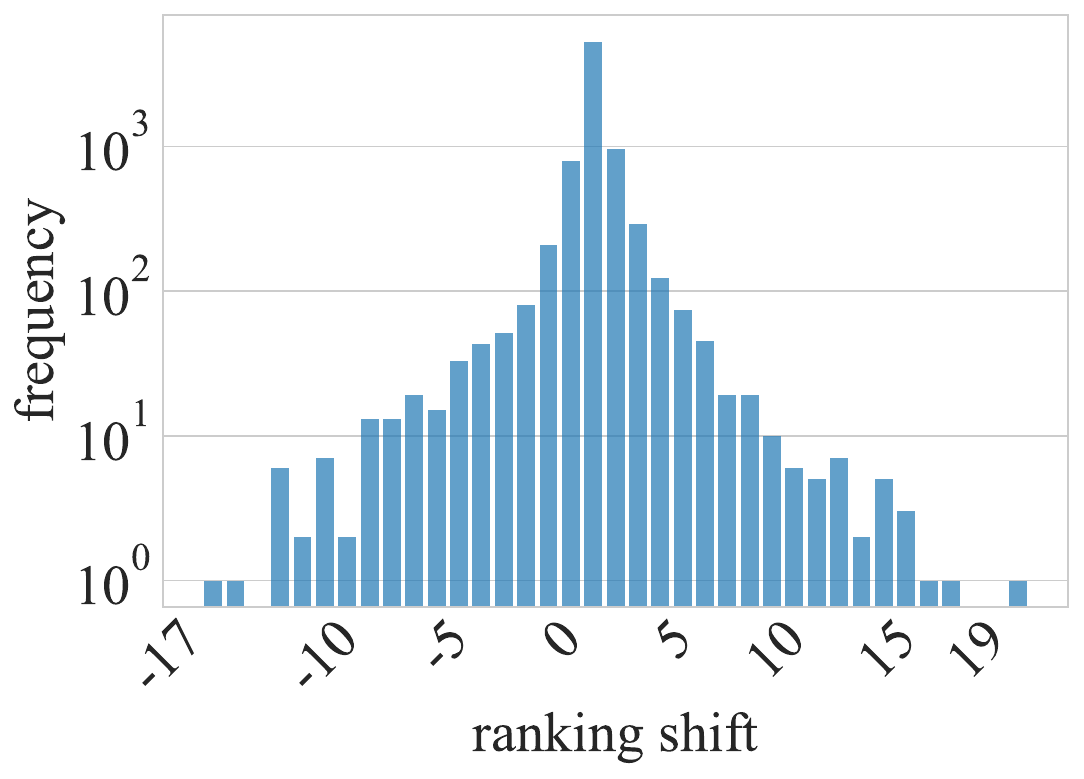}
       \caption{10 equi-depth buckets rank shifts}
   \end{subfigure}
   \caption{Frequencies of rank shifts of the \emph{age} feature when using (a) 5 equi-width, (b) 5 equi-depth, (c) 10 equi-width, or (d) 10 equi-depth buckets on ACS Income. The frequencies are shown on a log scale for better readability. Negative rank shifts represent ``demotion'' of \emph{age} where the rank values increase (e.g., from Rank 1 to 10) making \emph{age} less important, while positive shifts represent rank ``promotion'' where the importance of \emph{age} increases. 
   }\label{fig:bucketization}
\end{figure*}

Figure~\ref{fig:bucketization} shows the frequencies of rank changes of the \emph{age} feature when using 5 or 10 equi-width or equi-depth buckets on ACS Income. For all scenarios, we observe that bucketization can have drastic impacts on a non-trivial portion of individuals, and that the importance of \emph{age} may change by as much as 20 positions in the ranking.

\begin{figure*}[t]
    \begin{subfigure}[t]{0.32\textwidth}
        \includegraphics[width=\linewidth]{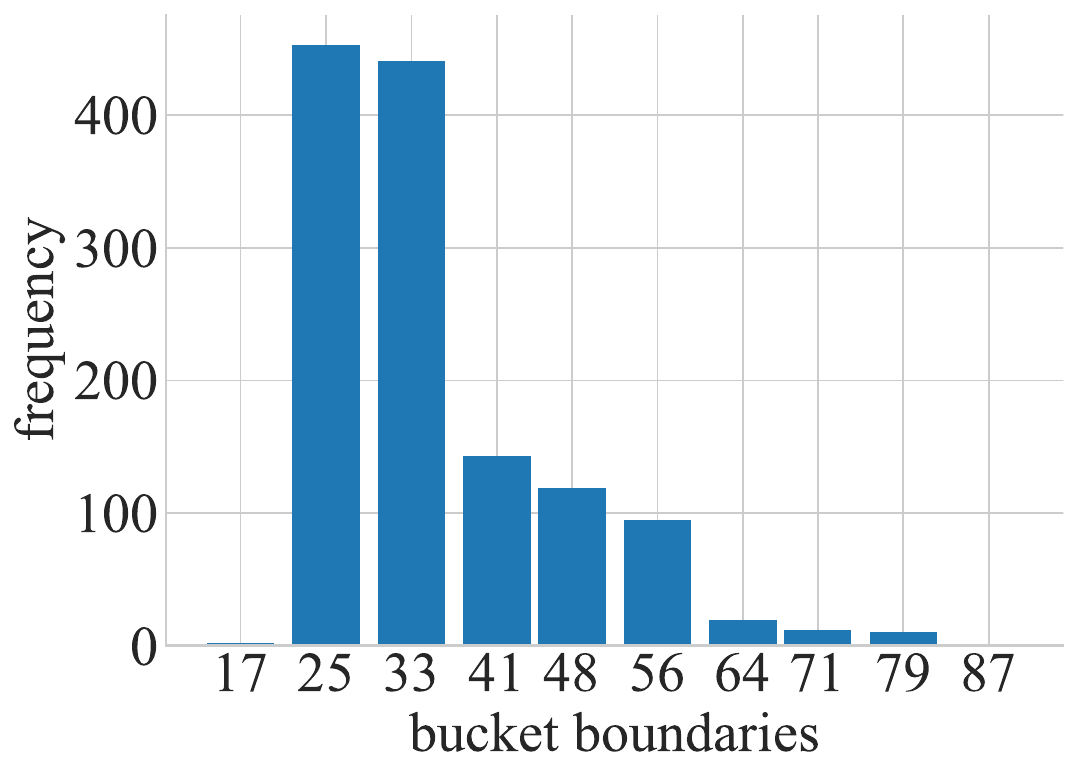}
        \caption{demotions of the rank of \emph{age}}
        \label{fig:sensitivity1b:a}
    \end{subfigure} 
    \begin{subfigure}[t]{0.32\textwidth}
        \includegraphics[width=\linewidth]{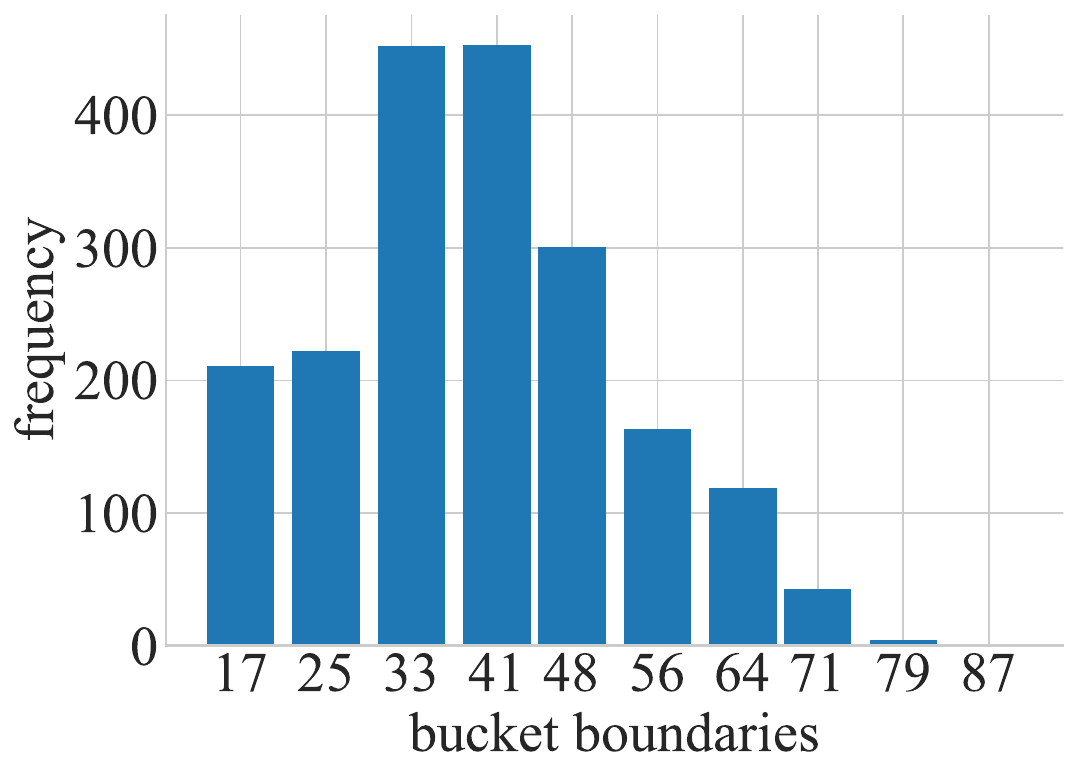}
        \caption{promotions of the rank of \emph{age}}
        \label{fig:sensitivity1b:b}
    \end{subfigure} 
    \begin{subfigure}[t]{0.32\textwidth}
        \includegraphics[width=\linewidth]{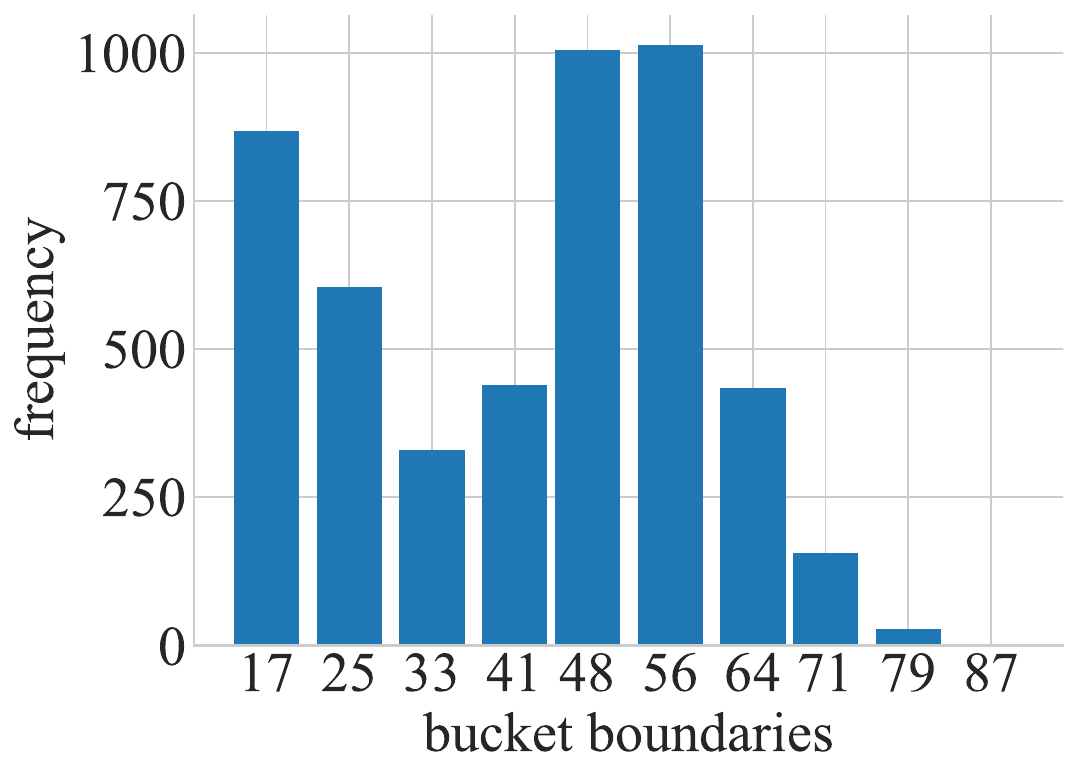}
        \caption{the rank of \emph{age} remains unchanged}
        \label{fig:sensitivity1b:c}
    \end{subfigure}  

    \caption{
    Frequency plot for varying numbers of equi-width buckets on \emph{age} on ACS Income. Bucket boundaries are shown on the  $x$-axis.
    }
    
    \label{fig:sensitivity1b} 
\end{figure*}

In Figure~\ref{fig:sensitivity1b}, we vary the number of equi-width buckets and show the number of observations for which the rank of the \emph{age} feature increased, decreased, or did not change. We also perform the same experiment for individuals for whom \emph{age} is the highest-ranked (\ie most important) feature, see Figure~\ref{fig:sensitivity1b-appendix} in the Appendix. We can see in Figures~\ref{fig:sensitivity1b:a} and~\ref{fig:sensitivity1b:b} that  the third bucket has the highest volatility. This demonstrates that bucketization can influence specific demographics more than others.


We also investigate how the SHAP ranking sensitivity changes based on the confusion matrix position of data points, \ie whether an individual was a true positive (TP), false positive (FP), true negative (TN), or false negative (FN). This has important implications for fairness, as many fairness metrics are based on metrics derived from the confusion matrix of a model~\cite{saleiro2018aequitas}. Interestingly, we find that changes are not uniform across these groups, as seen in Figure~\ref{fig:subgroups} in the Appendix, 
which quantifies the decrease in the importance of \emph{age} relative to other features as the number of histogram buckets increases.  Depending on how much the adversary intends to lower the importance of a protected feature, the adversary can choose a specific number of buckets. For example, suppose that an adversary is using ACS Income and wants to make \emph{age} look unimportant for false-negative observations using the data in Figure~\ref{fig:subgroups}. Looking at Figure~\ref{fig:subgroups:d}, the adversary may decide to use 4 or 7 buckets where a large portion of the observations consider \emph{age} to be unimportant (``rank 5 or below'').


\textbf{In summary,} we showed that SHAP is highly sensitive to bucketization, with the relative importance of \emph{age} changing by as much as 20 rank positions in some cases. Furthermore, when \emph{age} is the most important feature, its importance frequently drops by 3--5 positions in the ranking.

\subsection{Categorical features (\emph{race})}

We also evaluate SHAP's sensitivity to the representation of the categorical feature \emph{race}. We apply two preprocessing strategies: one-vs-rest (\textsf{OvR}) and a combinatorial merging approach. In \textsf{OvR}, each \emph{race} value is isolated while the remaining values are grouped, and a classifier is trained on the modified feature. The rest of the evaluation follows the procedure described earlier. The different merging strategies are shown and further described in Table~\ref{tbl:buckets}.


\begin{table}[t]
  \caption{Bucketization strategies for the \emph{race} feature. \textsc{Base} retains the original categories. \textsc{OvR} (``one vs. rest'') creates two categories: one for the specified value and one for all others. The \textsc{2 buckets} and \textsc{3 buckets} strategies group values into 2 or 3 categories, respectively. Each row in the table corresponds to a strategy; buckets are separated by commas, and merged values are indicated with a plus sign (+).}
  \label{tbl:buckets}
  \centering
  \begin{tabular}{cc}
    \toprule
  Strategy & Buckets \\
    \midrule
    \textsc{Base} & White, Black, Asian, Other \\
    \midrule
    \multirow{4}{*}{One vs. rest (\textsc{OvR})} & White, Rest \\
            & Black, Rest \\
            & Asian, Rest \\
            & Other, Rest \\
    \midrule
    \multirow{4}{*}{\textsc{2 buckets}} & White, Black + Asian + Other \\
     & White + Black , Asian + Other 
     \\
     & White + Asian, Black + Other 
     \\
     & White, Other \\
    \midrule
    \multirow{2}{*}{\textsc{3 buckets}} & White, Black, Asian + Other \\
     & White, Asian, Black + Other \\
    \bottomrule
  \end{tabular}
\end{table}

\begin{figure*}[t]
    \begin{subfigure}[t]{0.33\textwidth}
        \includegraphics[width=\linewidth]{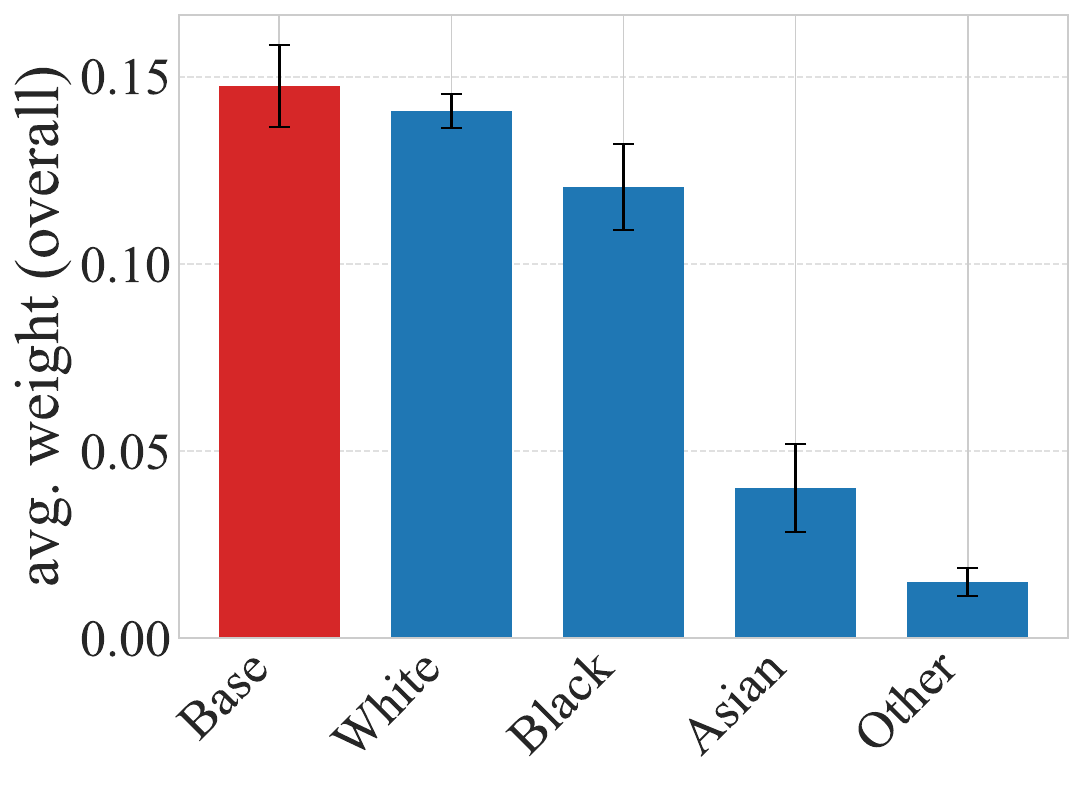}
        \caption{}
        \label{fig:avg-shap-race}
    \end{subfigure}
    \begin{subfigure}[t]{0.33\textwidth}
        \includegraphics[width=\linewidth]{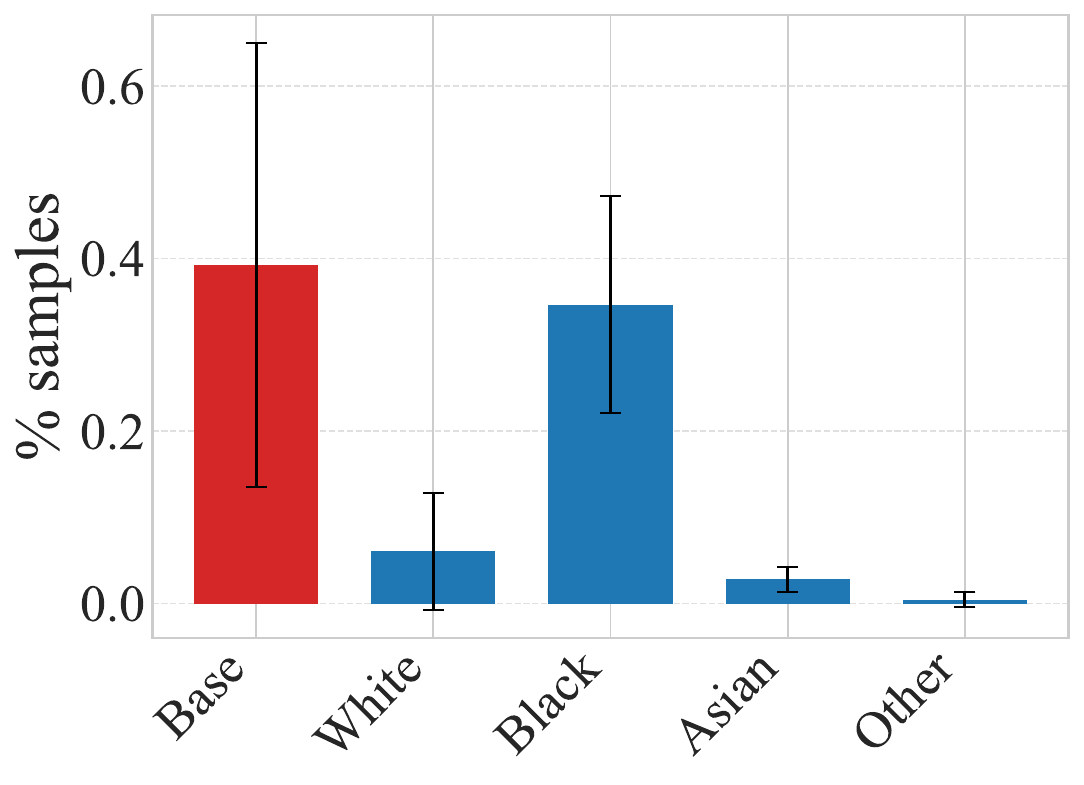}
        \caption{}
        \label{fig:top1freq-race}
    \end{subfigure}
    \begin{subfigure}[t]{0.33\textwidth}
        \includegraphics[width=\linewidth]{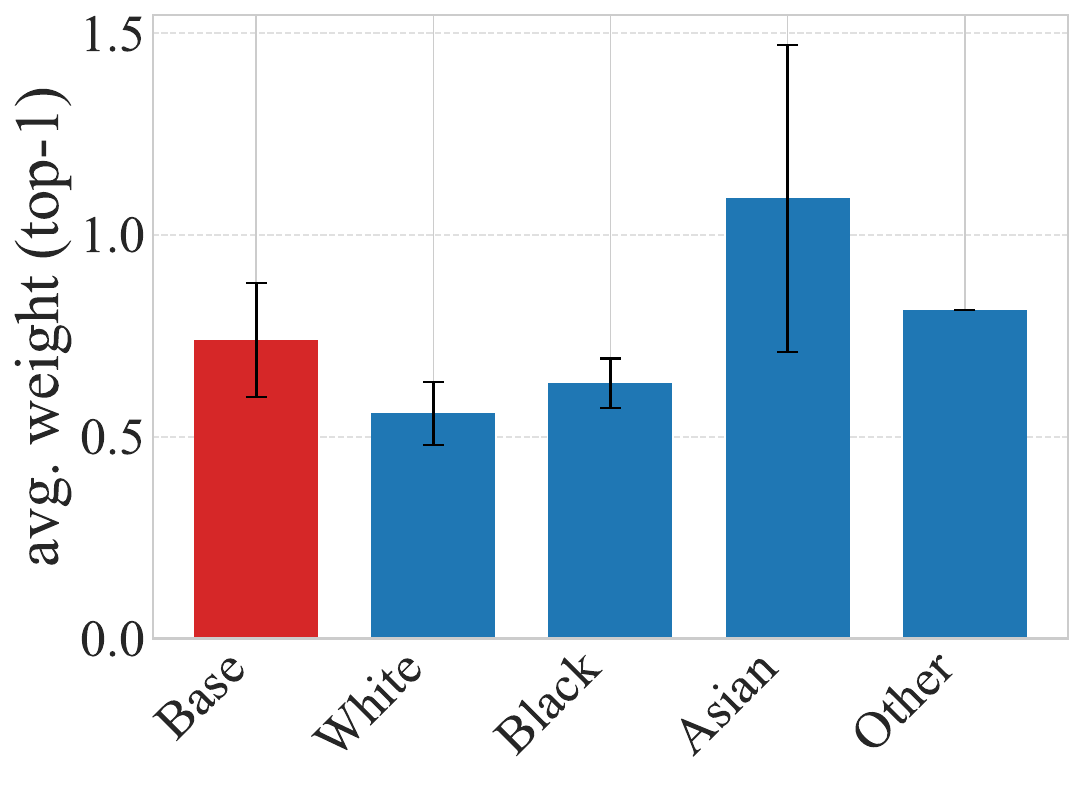}
        \caption{}
        \label{fig:top-shapley-race}
    \end{subfigure} 
    \caption{Effect on SHAP values using \textsc{OvR} bucketization for the \emph{race} feature using the ACS Income dataset. We use each feature value (White, Asian, Black, or Other) as a category and assign the remaining values to a single category. (a) Average SHAP values of the \emph{race} feature. (b) \textsc{OvR} setting versus the portion of observations where \emph{race} is the most important feature. (c) \textsc{OvR} setting versus the mean SHAP value of \emph{race} for the observations where \emph{race} is the most important feature.}\label{fig:sensitivity2}
\end{figure*}

Figures~\ref{fig:sensitivity2} (a)--(c) show how bucketization affects the \emph{race} feature in the \textsf{OvR} case. In particular, we compare several settings: \textsf{Base} (no merging), White vs. others, Asian vs. others, Black vs. others, and Non-(White, Asian, Black) vs. others. Overall, the average SHAP value of the \emph{race} feature decreases compared to \textsf{Base} for all settings as shown in Figure~\ref{fig:avg-shap-race}. Even if the decrease seems minor for the White-versus-Rest strategy, the fraction of samples that have \emph{race} as their most important feature drastically decreases, as shown in Figure~\ref{fig:top1freq-race}, demonstrating the effectiveness of bucketization. For Asian-versus-Rest, both of these values decrease significantly. We perform additional analyses in Figure~\ref{fig:top-shapley-race}, which shows the average SHAP value of \emph{race} only for observations with \emph{race} as the most important feature. For Asian-versus-Rest, the average SHAP value is higher than that of \textsf{Base}, which means that these (few) observations are more likely to be discriminated against based on \emph{race}.



\textbf{In summary,} we showed that SHAP is highly sensitive to bucketization for categorical features where the bucketization effectively shifts the importance from race to other features.

\section{Second contribution: A feature engineering attack on SHAP}
\label{sec:attack}

\subsection{Deliberate manipulation of SHAP values}
\label{sec:manipulation} 

In Section~\ref{sec:sensitivity}, we demonstrated that post-hoc SHAP explanations are sensitive to the way features are encoded. Importantly, this sensitivity can be exploited to intentionally manipulate feature importance reported by SHAP. 

\paragraph{Audit scenario.} Similar to ~\citet{DBLP:conf/iclr/LabergeA0MK23}, we consider a two-party audit scenario. The first party is a vendor that has full, white-box access to the data, the classifier model, and the data engineering and modeling pipelines. The vendor is able to make data engineering and modeling decisions and implement them into the respective pipelines. The second party is an auditor, who receives static copies of pre-processed data (\ie the data that is prepared for modeling) and the model, to which it has black-box access. While the auditor cannot perform any engineering, it can generate feature-based explanations of model predictions for any observation in the dataset using SHAP. This allows the auditor to inspect how often the model appears to use protected features (according to SHAP explanations) to make classification decisions. Notably, the vendor has an adversarial goal: it wants to build models that make use of protected features, but it does not want those features to appear to have high importance according to SHAP when inspected by an auditor.

\paragraph{Formalization.} The dataset $D$ consists of $|D|$ training examples as tuples $(\mathbf{x}_i, y_i)$, where example $i$ corresponds to the $i$-th individual in the dataset. Each $\mathbf{x} \in \mathbb{R}^n$, where $n$ is the number of features, and that individual's outcome (or target) is given by $y \in \{0,1\}$. We use $a$ to denote the feature at a specific index $x_a \in \mathbf{x}$, which represents the protected feature for each individual. The dataset $D$ is used to learn a machine learning classifier $f : \mathcal{X} \rightarrow [0,1]$.

Recall from the description of the audit scenario, that the vendor is able to make data engineering decisions and implement them before training the classifier $f$. Let $\mathcal{T}$ be the set of all possible feature transformations that could be applied to the protected feature $a$. Each transformation $\tau \in \mathcal{T}$ is a function of the form $\tau : \mathcal{A} \rightarrow \mathcal{S}$, where $\mathcal{A}$ is the set of possible values for the protected feature $a$, and $\mathcal{S}$ is the set of all possible values for the transformed feature. We refer to $\tau(a)$ as the transformation of feature $a$. Then generally, the manipulation framework we propose uses the following objective:
\begin{align}
    \min_{\mathcal{T}} -\text{SHAP\_Rank}(a,f,D),\text{~s.t.~} \lambda > \epsilon
\end{align}
where $\text{SHAP\_Rank}(a,f,D)$ is a function that returns the SHAP rank of feature $a$ under model $f$ and with data $D$, $\lambda$ is the \emph{fidelity} of the explanation (as defined in Section~\ref{sec:preliminaries:exp}), and $\epsilon$ is a user-defined threshold for fidelity. The purpose of the fidelity constraint is to ensure that the feature transformation remains faithful to the original model (and, consequently, to the original explainer).

We now describe a particular instantiation of this framework for manipulating continuous features via the bucketization feature transformation. Suppose that $a \in \mathbb{R}^{\geq 0}$ is a continuous feature. We can define the feature transformation $\tau_K: \mathbb{R}^{\geq 0} \rightarrow \{0,1, \dots, k \}$ that discretizes $a$ into $k$ buckets in the following way:
\begin{align}
\tau_K(a) &= \begin{cases}
        \text{0}, ~b_0 < a < b_1 \\
        \text{1}, ~b_1 < a < b_2 \\
        ~\vdots \\
        \text{K}, ~b_{k-1} < a < b_k \\
        \end{cases}
\end{align}
Where $b_0 < b_1 < \dots < b_k$ are the upper-lower bounds for each bucket. Based on the choices for the upper-lower bounds used to define the cases of $\tau_k$, applying the transformation $\tau_k(a)$ can be used to induce a particular set of $k$ partitions over the data $D$. Let $\mathcal{P}_k$ be the set of all $k$-partitions over $D$. Note that this occurs upstream in our machine learning pipeline, so $f$ will always be trained on data with the transformed feature. Then our objective is:
\begin{align}
\min_{k \in \{0,1, \dots,k\}, P \in \mathcal{P_k}} -\text{SHAP\_rank}(a,f,D), \text{~s.t.~} b_0 < \dots < b_k, \lambda > \epsilon
\label{eq:bob_objective}
\end{align}


We can solve this problem using Bayesian Optimization~\cite{DBLP:conf/ifip7/Mockus74}, which is commonly used to solve black-box optimization problems by constructing a posterior distribution of Gaussian functions that best describe the unknown function to optimize. Bayesian Optimization is particularly appropriate for this problem because it is effective when the objective function is expensive to evaluate, as is often the case with the function $\text{SHAP\_rank}$.

\subsection{Bucketization attack experiments}
\label{sec:bayes_opt_experiments}

We now perform the data engineering attacks described in the previous section, which use Bayesian Optimization (BO) to tune bucket boundaries when using \emph{age}. 
Compared to equi-width bucketization, we observe sharper changes in the SHAP rank of \emph{age}, but also a more substantial decrease in fidelity. (Although not shown here, the comparison with equi-depth bucketization is very similar.)

The experimental results in this section are computed using 5-fold cross-validation, with metrics averaged across folds. We report results for both ACS Income and ACS Public Coverage.

For BO, we tune four bucket boundaries between fixed minimum and maximum values. Using \emph{age}, we set the minimum and maximum values to be 17 and 94 years old, respectively. We then perform 300 iterations where the objective function is the same as Equation (\ref{eq:bob_objective}). For the fidelity constraint, we require that the fidelity is at least as good as that in the equi-width setting. As a result, the BO attack significantly outperforms \textsf{Base} (no bucketization) and mostly outperforms the equi-width setting. We conclude that BO thus may be used to hide the contributions of \emph{age} on model predictions. 

\begin{figure}[t]
    \begin{subfigure}[t]{0.4\textwidth}
        \includegraphics[width=\linewidth]{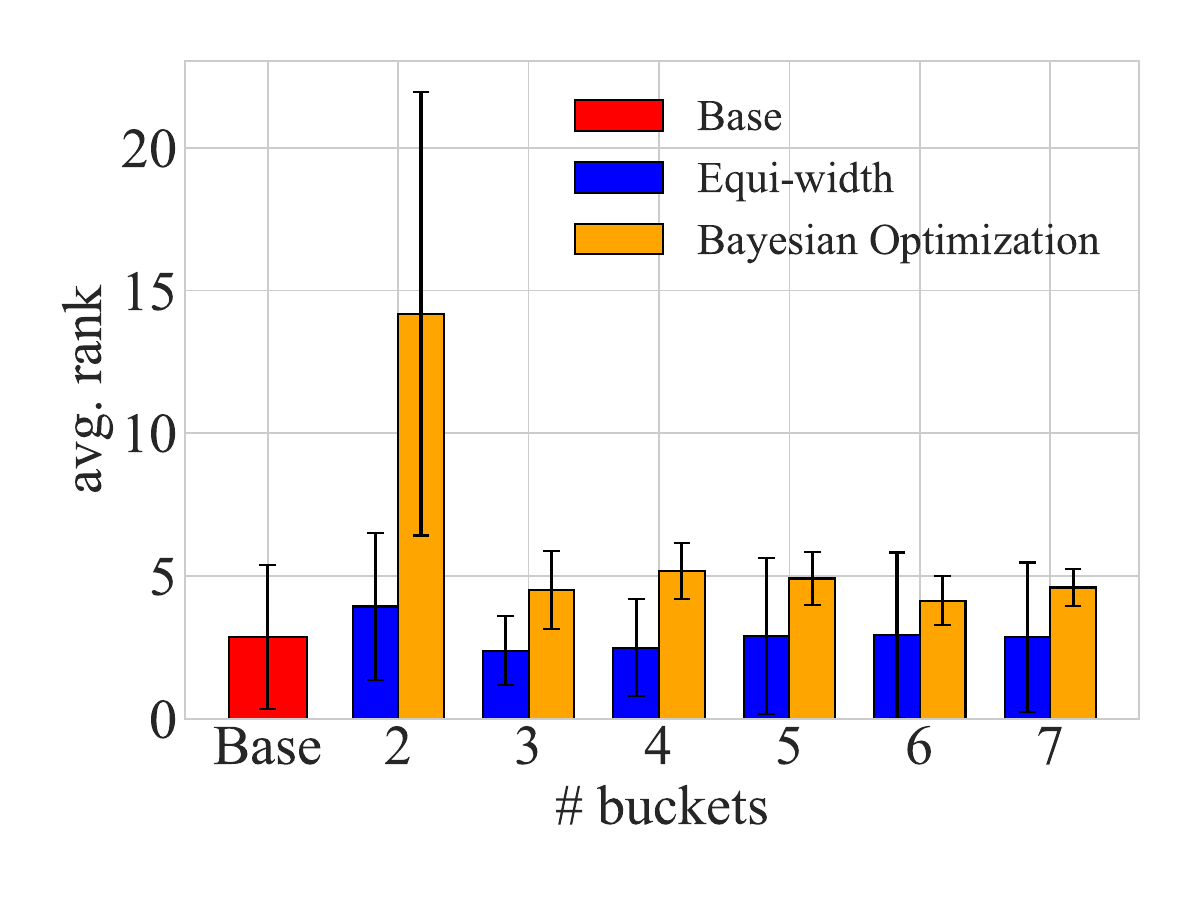}
        \caption{ACS Income ranks}
    \end{subfigure}
    \begin{subfigure}[t]{0.4\textwidth}
        \includegraphics[width=\linewidth]{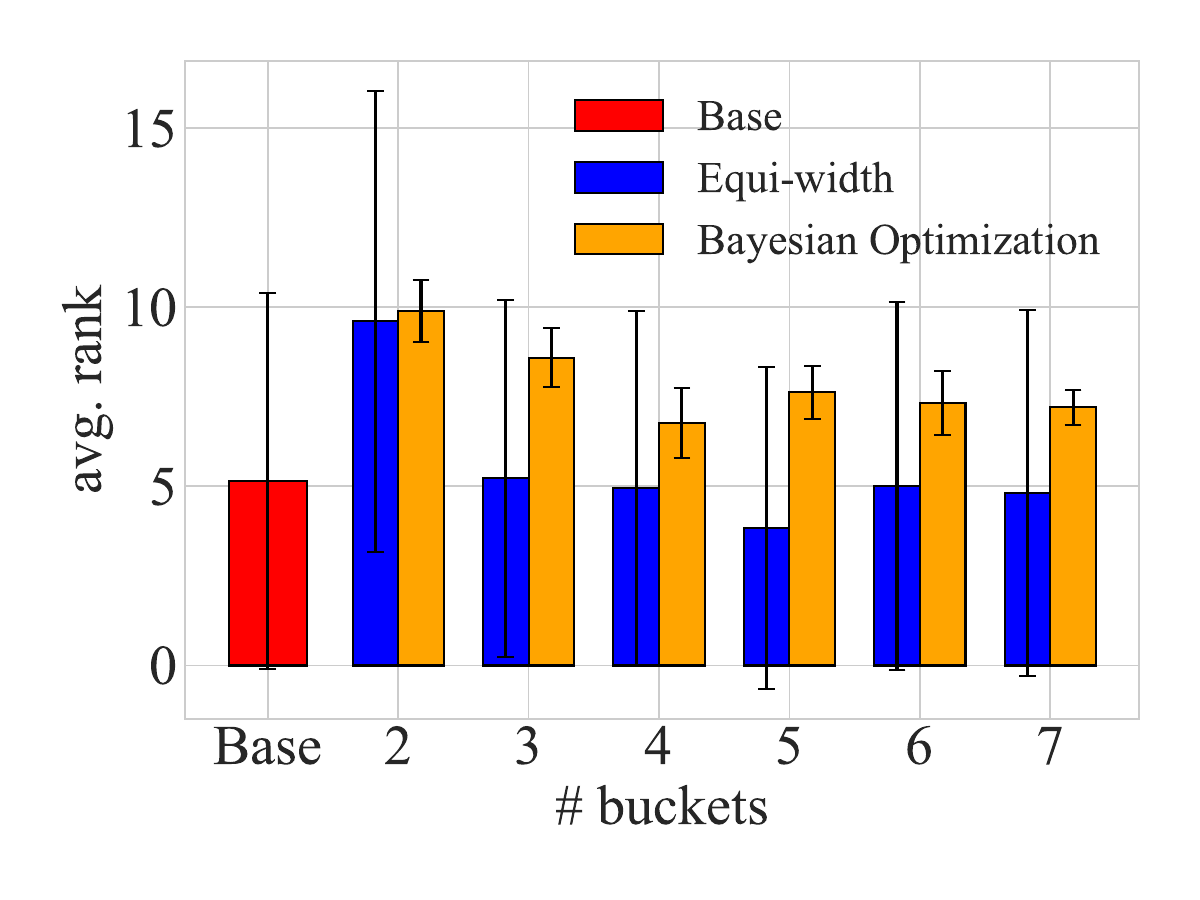}
        \caption{ACS Public Coverage ranks}
    \end{subfigure} 
    \caption{SHAP value rank trends of the \emph{age} feature on ACS Income and ACS Public Coverage.  Higher average rank means that feature has lower importance.  Fidelity under attack is at least as high as under equi-width bucketization in each case. 
    }\label{fig:attackbucketcount}
\end{figure}

Figure~\ref{fig:attackbucketcount} shows how the importance rank of \emph{age}  changes when varying the number of buckets (and thus the number of BO parameters) using ACS Income and Public Coverage. As the number of buckets increases, our attack consistently shows comparable or higher ranks compared to using the \textsf{Base} average rank, although the gap decreases. The gap decrease is due to the difficulty in solving high-dimensional BO problems.  At the same time, the constraint on fidelity ensures that our attack is always at least as meaningful as the equi-width bucketization. 
Table~\ref{tab:age_fidelity} shows the fidelity values of the attacks on \emph{age}, with bucketization of Figure~\ref{fig:attackbucketcount}. Even when the average rank of \emph{age} drops dramatically, the fidelity values are at least 88\%.


\begin{table}[t]
    \centering
    \caption{Fidelity values of bucketizations on \emph{age} for ACS Income and ACS Public Coverage.}
    \begin{tabular}{lcc}
        \toprule
        \textbf{Buckets} & \textbf{ACS Income} & \textbf{ACS Public Coverage} \\
        \midrule
        2 & 88.60 $\pm$ 0.09 & 94.38 $\pm$ 0.83 \\
        3 & 91.37 $\pm$ 0.26 & 95.99 $\pm$ 0.13 \\
        4 & 92.39 $\pm$ 0.21 & 96.95 $\pm$ 0.03 \\
        5 & 93.15 $\pm$ 0.13 & 96.85 $\pm$ 0.04 \\
        6 & 94.67 $\pm$ 0.08 & 97.25 $\pm$ 0.03 \\
        7 & 94.59 $\pm$ 0.04 & 97.43 $\pm$ 0.03 \\
        \bottomrule
    \end{tabular}
    \label{tab:age_fidelity}
\end{table}

\begin{figure}[t]
    \begin{subfigure}[t]{0.35\textwidth}
        \includegraphics[width=\linewidth]{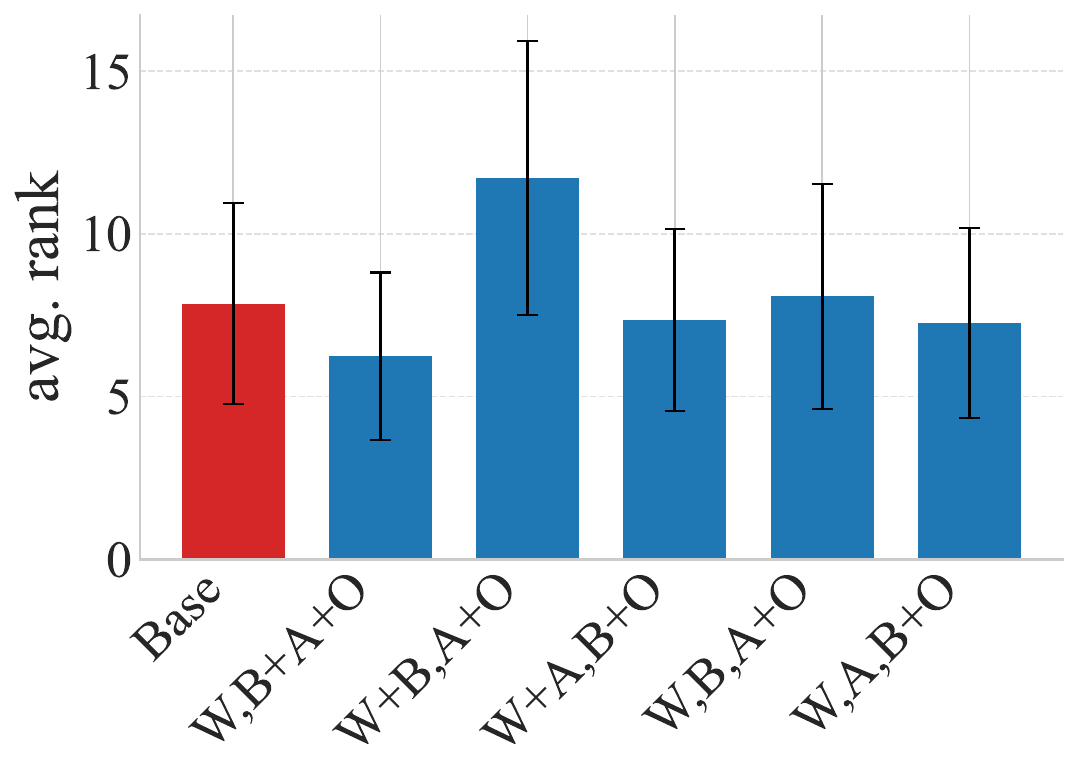}
        \caption{ACS Income}
    \end{subfigure}
    \begin{subfigure}[t]{0.35\textwidth}
        \includegraphics[width=\linewidth]{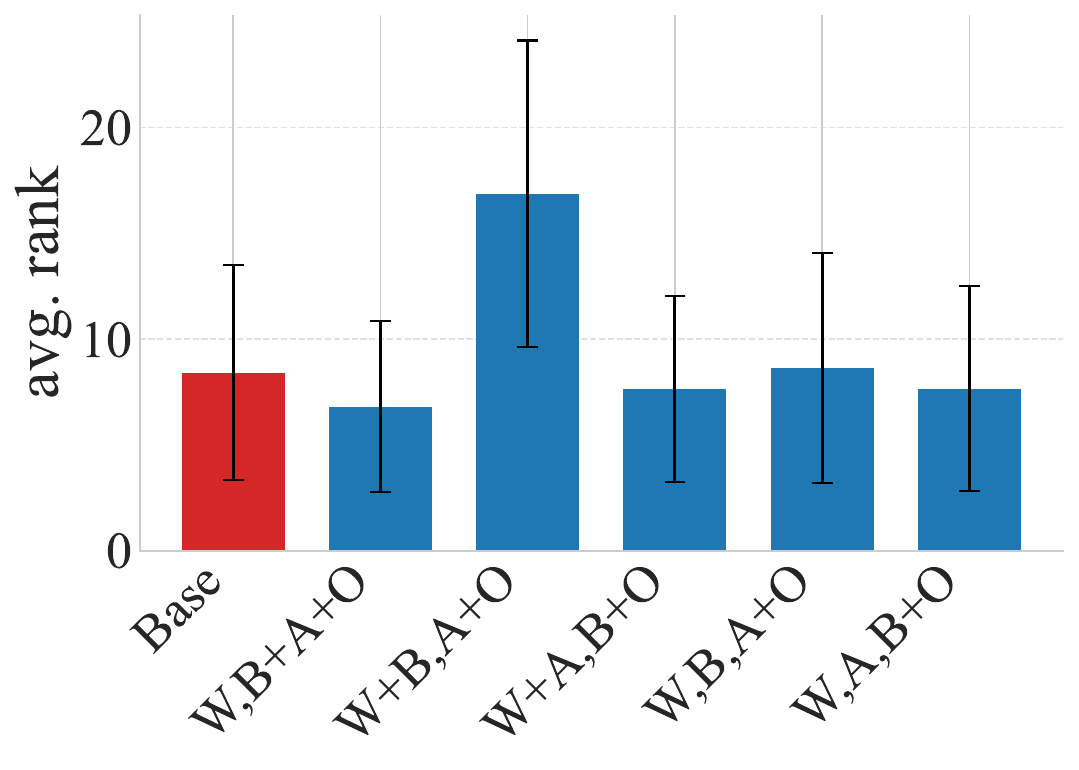}
        \caption{ACS Public Coverage}
    \end{subfigure} 
    \caption{SHAP value rank trends of the \emph{race} feature using the ACS Income and ACS Public Coverage datasets. Simply combining two races into a single bucket dramatically increases the average SHAP value rankings.
    }\label{fig:attackvaryingbuckets}
\end{figure}

When using \emph{race}, BO cannot be performed on categorical values, so we instead combine races for the bucketization attack as in Table~\ref{tbl:buckets}, but consider more combinations. Figure~\ref{fig:attackvaryingbuckets} shows the how the rank of \emph{race} changes on ACS Income and ACS Public Coverage. For the \{White+Black, Asian+Other\} bucketization on both tasks, the rank value of \emph{race} is significantly higher (\ie \emph{race} is far less important) compared to the \textsf{Base} strategy.  We suspect that  \emph{race=White} by itself is a strong signal when making predictions, but when combined with another race into the same bucket, its importance is diluted.
Table~\ref{tab:race_fidelity} shows the fidelity values of the attacks on \emph{race} for the bucketizations of Figure~\ref{fig:attackvaryingbuckets}. At least 98\% of all explanations have perfect fidelity (\ie reconstruct the true outcome).

\begin{table}[t]
    \centering
    \caption{Fidelity value of bucketizations on \emph{race} for ACS Income and ACS Public Coverage.}
    \begin{tabular}{lcc}
        \toprule
        \textbf{Buckets} & \textbf{ACS Income} & \textbf{ACS Public Coverage} \\
        \midrule
        W, B+A+O & 98.56 $\pm$ 0.48 & 99.57 $\pm$ 0.09 \\
        W+B, A+O & 98.10 $\pm$ 0.10 & 98.00 $\pm$ 0.19 \\
        W+A, B+O & 99.16 $\pm$ 0.09 & 99.65 $\pm$ 0.10 \\
        W, B, A+O & 99.72 $\pm$ 0.01 & 99.67 $\pm$ 0.10 \\
        W, A, B+O & 99.65 $\pm$ 0.05 & 99.74 $\pm$ 0.09 \\
        \bottomrule
    \end{tabular}
    \label{tab:race_fidelity}
\end{table}

\textbf{In summary,} we showed that, on ACM Income and Public Coverage, our BO attack mostly outperforms equi-width bucketization in terms of increasing the SHAP ranks of \emph{age} without compromising fidelity. Our attack is also effective when using \emph{race} where we combine races for bucketization. In particular, on ACS Public Coverage, combining White and Black individuals into a single category brings the importance of \emph{race} close to zero.

\section{Discussion}
\label{sec:discussion}

We note that any sociotechnical system that incorporates AI/ML is not a monolith: it impacts many stakeholders, including practitioners, compliance officers, auditors, affected individuals, and society at large. In this section, we state the implications of our work for practitioners and auditors seeking to ensure the responsible implementation of AI/ML systems.

Our experimental results show that SHAP, one of the most widely-used post-hoc explanation methods, is highly sensitive to upstream data engineering decisions. Furthermore, we demonstrate that a principled approach using Bayesian Optimization can exploit this sensitivity to manipulate SHAP. Below, we discuss the practical implications of this finding.

\paragraph{Practitioners.} Our results further highlight the importance of thoughtful feature engineering when building machine learning pipelines, and of assessing how those decisions may impact the accuracy, fairness, and explainability of the systems being built and deployed~\cite{DBLP:journals/cacm/StoyanovichAHJS22}. \citet{DBLP:journals/kais/KamiranC11} and~\citet{DBLP:journals/jmlr/ZafarVGG19} showed that varying feature representations can affect the fairness of a machine learning classifier. We expand on this result, showing that these effects extend to model explainability—particularly when using the post-hoc explanation method SHAP. Notably, this recommendation can also be framed as an opportunity. In line with earlier framing by~\citet{NMI_2020},~\citet{DBLP:conf/fat/BellSNS22} showed that the explainability of a system depends on the context of use, the data, the underlying model type, the stakeholders, and the specific questions a human is trying to answer about the system. Practitioners could use these factors to inform how they encode features to \emph{improve} the explainability of their pipelines.

\paragraph{Auditors.} In this work, we showed how seemingly innocuous decisions---such as the bucketization of the \emph{age} feature---can be used to manipulate the SHAP explainer. Our experiments demonstrate that one can successfully reduce a feature’s apparent importance while \emph{minimally impacting} the explanation’s fidelity to the original model's outcomes. Importantly, such manipulation could be exploited by adversarial actors to obscure discrimination in their models. While developing a technical defense is beyond the scope of this paper, we offer the following recommendation to auditors: governance and audit frameworks for machine learning systems should be expanded to account for data engineering decisions.

\paragraph{Limitations.} 
This work has several limitations. While we do perform an attack on categorical features, it relies on predefined, semantically meaningful groupings. More work is needed to develop methods for semi-automatically exploring the combinatorial space of groupings, particularly for high-cardinality features. A naïve approach---exhaustively enumerating all bucketings to find the one that most impacts the importance rank of \emph{race} while preserving model fidelity---is computationally infeasible. Future work should develop principled techniques for navigating this space efficiently.

Another limitation is that we consider sensitive features in isolation. Our experiments explore the sensitivity of explanations to feature representations for \emph{age} and \emph{race} separately, with targeted attacks designed independently for each. In practice, sensitive attributes often interact---and their combined effects can influence model behavior and interpretability. Future work should develop holistic methods that jointly consider multiple sensitive features for explainability and robustness.

\section{Conclusions and future work}
\label{sec:conclusion}

We explored how common data engineering techniques affect local feature-based explanations from methods like SHAP, and showed that subtle preprocessing choices can significantly alter explanations. We also introduced a feature engineering attack that hides the importance of protected features with minimal impact on predictions. Finally, we called for more robust explanation frameworks that consider not only accuracy and fairness, but also the influence of data engineering---especially in transparency-critical settings.

While our work highlighted SHAP's sensitivity to common preprocessing operations and their potential for misuse, the same insights can be applied constructively. Instead of designing attacks, similar techniques can inform data engineering choices that improve the explainability and robustness of classification decisions.

For example, identifying preprocessing choices that consistently elevate the importance of semantically meaningful features may lead to models that better reflect human reasoning and are easier to audit. These insights could also support the development of classifiers with greater control over the distribution of feature importance. Future work should extend this approach to sets of features---continuous, categorical, or mixed---and investigate how thoughtful data engineering can help align model explanations with stakeholder expectations. Finally, future research should examine the impact of more complex data engineering transformations on SHAP explanations.


\begin{acks}
This work was supported in part by the NYU-KAIST Partnership and by the Institute of Information \& Communications Technology Planning \& Evaluation (IITP) with a grant funded by the Ministry of Science and ICT (MSIT) of the Republic of Korea in connection with the Global AI Frontier Lab International Collaborative Research. (No. RS-2024-00509258 and No. RS-2024-00469482). This work was also supported in part by the National Research Foundation of Korea (NRF) grant funded by the Korea government (MSIT) (No.\@ NRF-2022R1A2C2004382), the US National Science Foundation (NSF) Awards No. 2312930 and 2326193, and by NSF GRFP DGE-2234660.
\end{acks}

\bibliographystyle{ACM-Reference-Format}


\begin{thebibliography}{39}


\ifx \showCODEN    \undefined \def \showCODEN     #1{\unskip}     \fi
\ifx \showDOI      \undefined \def \showDOI       #1{#1}\fi
\ifx \showISBNx    \undefined \def \showISBNx     #1{\unskip}     \fi
\ifx \showISBNxiii \undefined \def \showISBNxiii  #1{\unskip}     \fi
\ifx \showISSN     \undefined \def \showISSN      #1{\unskip}     \fi
\ifx \showLCCN     \undefined \def \showLCCN      #1{\unskip}     \fi
\ifx \shownote     \undefined \def \shownote      #1{#1}          \fi
\ifx \showarticletitle \undefined \def \showarticletitle #1{#1}   \fi
\ifx \showURL      \undefined \def \showURL       {\relax}        \fi
\providecommand\bibfield[2]{#2}
\providecommand\bibinfo[2]{#2}
\providecommand\natexlab[1]{#1}
\providecommand\showeprint[2][]{arXiv:#2}

\bibitem[Alwarthan et~al\mbox{.}(2022)]%
        {alwarthan2022explainable}
\bibfield{author}{\bibinfo{person}{Sarah Alwarthan}, \bibinfo{person}{Nida Aslam}, {and} \bibinfo{person}{Irfan~Ullah Khan}.} \bibinfo{year}{2022}\natexlab{}.
\newblock \showarticletitle{An explainable model for identifying at-risk student at higher education}.
\newblock \bibinfo{journal}{\emph{IEEE Access}}  \bibinfo{volume}{10} (\bibinfo{year}{2022}), \bibinfo{pages}{107649--107668}.
\newblock


\bibitem[Baniecki and Biecek(2022)]%
        {DBLP:conf/aaai/BanieckiB22}
\bibfield{author}{\bibinfo{person}{Hubert Baniecki} {and} \bibinfo{person}{Przemyslaw Biecek}.} \bibinfo{year}{2022}\natexlab{}.
\newblock \showarticletitle{Manipulating {SHAP} via Adversarial Data Perturbations (Student Abstract)}. In \bibinfo{booktitle}{\emph{Thirty-Sixth {AAAI} Conference on Artificial Intelligence, {AAAI} 2022, Thirty-Fourth Conference on Innovative Applications of Artificial Intelligence, {IAAI} 2022, The Twelveth Symposium on Educational Advances in Artificial Intelligence, {EAAI} 2022 Virtual Event, February 22 - March 1, 2022}}. \bibinfo{publisher}{{AAAI} Press}, \bibinfo{pages}{12907--12908}.
\newblock


\bibitem[Barocas et~al\mbox{.}(2020)]%
        {DBLP:conf/fat/BarocasSR20}
\bibfield{author}{\bibinfo{person}{Solon Barocas}, \bibinfo{person}{Andrew~D. Selbst}, {and} \bibinfo{person}{Manish Raghavan}.} \bibinfo{year}{2020}\natexlab{}.
\newblock \showarticletitle{The hidden assumptions behind counterfactual explanations and principal reasons}. In \bibinfo{booktitle}{\emph{FAT* '20: Conference on Fairness, Accountability, and Transparency, Barcelona, Spain, January 27-30, 2020}}. \bibinfo{publisher}{{ACM}}, \bibinfo{pages}{80--89}.
\newblock


\bibitem[Bell et~al\mbox{.}(2022)]%
        {DBLP:conf/fat/BellSNS22}
\bibfield{author}{\bibinfo{person}{Andrew Bell}, \bibinfo{person}{Ian Solano{-}Kamaiko}, \bibinfo{person}{Oded Nov}, {and} \bibinfo{person}{Julia Stoyanovich}.} \bibinfo{year}{2022}\natexlab{}.
\newblock \showarticletitle{It's Just Not That Simple: An Empirical Study of the Accuracy-Explainability Trade-off in Machine Learning for Public Policy}. In \bibinfo{booktitle}{\emph{FAccT '22: 2022 {ACM} Conference on Fairness, Accountability, and Transparency, Seoul, Republic of Korea, June 21 - 24, 2022}}. \bibinfo{publisher}{{ACM}}, \bibinfo{pages}{248--266}.
\newblock


\bibitem[Belle and Papantonis(2021)]%
        {belle2021principles}
\bibfield{author}{\bibinfo{person}{Vaishak Belle} {and} \bibinfo{person}{Ioannis Papantonis}.} \bibinfo{year}{2021}\natexlab{}.
\newblock \showarticletitle{Principles and practice of explainable machine learning}.
\newblock \bibinfo{journal}{\emph{Frontiers in big Data}}  \bibinfo{volume}{4} (\bibinfo{year}{2021}), \bibinfo{pages}{688969}.
\newblock


\bibitem[Benthall and Haynes(2019)]%
        {benthall2019racial}
\bibfield{author}{\bibinfo{person}{Sebastian Benthall} {and} \bibinfo{person}{Bruce~D Haynes}.} \bibinfo{year}{2019}\natexlab{}.
\newblock \showarticletitle{Racial categories in machine learning}. In \bibinfo{booktitle}{\emph{Proceedings of the conference on fairness, accountability, and transparency}}. \bibinfo{pages}{289--298}.
\newblock


\bibitem[Bhatt et~al\mbox{.}(2020)]%
        {bhatt2020explainable}
\bibfield{author}{\bibinfo{person}{Umang Bhatt}, \bibinfo{person}{Alice Xiang}, \bibinfo{person}{Shubham Sharma}, \bibinfo{person}{Adrian Weller}, \bibinfo{person}{Ankur Taly}, \bibinfo{person}{Yunhan Jia}, \bibinfo{person}{Joydeep Ghosh}, \bibinfo{person}{Ruchir Puri}, \bibinfo{person}{Jos{\'e}~MF Moura}, {and} \bibinfo{person}{Peter Eckersley}.} \bibinfo{year}{2020}\natexlab{}.
\newblock \showarticletitle{Explainable machine learning in deployment}. In \bibinfo{booktitle}{\emph{Proceedings of the 2020 conference on fairness, accountability, and transparency}}. \bibinfo{pages}{648--657}.
\newblock


\bibitem[Black et~al\mbox{.}(2022)]%
        {DBLP:conf/fat/BlackRB22}
\bibfield{author}{\bibinfo{person}{Emily Black}, \bibinfo{person}{Manish Raghavan}, {and} \bibinfo{person}{Solon Barocas}.} \bibinfo{year}{2022}\natexlab{}.
\newblock \showarticletitle{Model Multiplicity: Opportunities, Concerns, and Solutions}. In \bibinfo{booktitle}{\emph{FAccT '22: 2022 {ACM} Conference on Fairness, Accountability, and Transparency, Seoul, Republic of Korea, June 21 - 24, 2022}}. \bibinfo{publisher}{{ACM}}, \bibinfo{pages}{850--863}.
\newblock


\bibitem[Breiman(2001)]%
        {breiman2001statistical}
\bibfield{author}{\bibinfo{person}{Leo Breiman}.} \bibinfo{year}{2001}\natexlab{}.
\newblock \showarticletitle{Statistical modeling: The two cultures (with comments and a rejoinder by the author)}.
\newblock \bibinfo{journal}{\emph{Statist. Sci.}} \bibinfo{volume}{16}, \bibinfo{number}{3} (\bibinfo{year}{2001}), \bibinfo{pages}{199--231}.
\newblock


\bibitem[Chaddad et~al\mbox{.}(2023)]%
        {chaddad2023survey}
\bibfield{author}{\bibinfo{person}{Ahmad Chaddad}, \bibinfo{person}{Jihao Peng}, \bibinfo{person}{Jian Xu}, {and} \bibinfo{person}{Ahmed Bouridane}.} \bibinfo{year}{2023}\natexlab{}.
\newblock \showarticletitle{Survey of explainable AI techniques in healthcare}.
\newblock \bibinfo{journal}{\emph{Sensors}} \bibinfo{volume}{23}, \bibinfo{number}{2} (\bibinfo{year}{2023}), \bibinfo{pages}{634}.
\newblock


\bibitem[Datta et~al\mbox{.}(2016)]%
        {DBLP:conf/sp/DattaSZ16}
\bibfield{author}{\bibinfo{person}{Anupam Datta}, \bibinfo{person}{Shayak Sen}, {and} \bibinfo{person}{Yair Zick}.} \bibinfo{year}{2016}\natexlab{}.
\newblock \showarticletitle{Algorithmic Transparency via Quantitative Input Influence: Theory and Experiments with Learning Systems}. In \bibinfo{booktitle}{\emph{{IEEE} Symposium on Security and Privacy, {SP} 2016, San Jose, CA, USA, May 22-26, 2016}}. \bibinfo{publisher}{{IEEE} Computer Society}, \bibinfo{pages}{598--617}.
\newblock


\bibitem[Deck et~al\mbox{.}(2024)]%
        {DBLP:conf/fat/DeckSD024}
\bibfield{author}{\bibinfo{person}{Luca Deck}, \bibinfo{person}{Jakob Schoeffer}, \bibinfo{person}{Maria De{-}Arteaga}, {and} \bibinfo{person}{Niklas K{\"{u}}hl}.} \bibinfo{year}{2024}\natexlab{}.
\newblock \showarticletitle{A Critical Survey on Fairness Benefits of Explainable {AI}}. In \bibinfo{booktitle}{\emph{The 2024 {ACM} Conference on Fairness, Accountability, and Transparency, FAccT 2024, Rio de Janeiro, Brazil, June 3-6, 2024}}. \bibinfo{publisher}{{ACM}}, \bibinfo{pages}{1579--1595}.
\newblock


\bibitem[Ding et~al\mbox{.}(2021)]%
        {DBLP:conf/nips/DingHMS21}
\bibfield{author}{\bibinfo{person}{Frances Ding}, \bibinfo{person}{Moritz Hardt}, \bibinfo{person}{John Miller}, {and} \bibinfo{person}{Ludwig Schmidt}.} \bibinfo{year}{2021}\natexlab{}.
\newblock \showarticletitle{Retiring Adult: New Datasets for Fair Machine Learning}. In \bibinfo{booktitle}{\emph{Advances in Neural Information Processing Systems 34: Annual Conference on Neural Information Processing Systems 2021, NeurIPS 2021, December 6-14, 2021, virtual}}. \bibinfo{pages}{6478--6490}.
\newblock


\bibitem[Fresz et~al\mbox{.}(2024)]%
        {fresz2024should}
\bibfield{author}{\bibinfo{person}{Benjamin Fresz}, \bibinfo{person}{Elena Dubovitskaya}, \bibinfo{person}{Danilo Brajovic}, \bibinfo{person}{Marco~F Huber}, {and} \bibinfo{person}{Christian Horz}.} \bibinfo{year}{2024}\natexlab{}.
\newblock \showarticletitle{How should AI decisions be explained? Requirements for Explanations from the Perspective of European Law}. In \bibinfo{booktitle}{\emph{Proceedings of the AAAI/ACM Conference on AI, Ethics, and Society}}, Vol.~\bibinfo{volume}{7}. \bibinfo{pages}{438--450}.
\newblock


\bibitem[Hancox{-}Li(2020)]%
        {DBLP:conf/fat/Hancox-Li20}
\bibfield{author}{\bibinfo{person}{Leif Hancox{-}Li}.} \bibinfo{year}{2020}\natexlab{}.
\newblock \showarticletitle{Robustness in machine learning explanations: does it matter?}. In \bibinfo{booktitle}{\emph{FAT* '20: Conference on Fairness, Accountability, and Transparency, Barcelona, Spain, January 27-30, 2020}}. \bibinfo{publisher}{{ACM}}, \bibinfo{pages}{640--647}.
\newblock


\bibitem[Hancox{-}Li and Kumar(2021)]%
        {DBLP:conf/fat/Hancox-LiK21}
\bibfield{author}{\bibinfo{person}{Leif Hancox{-}Li} {and} \bibinfo{person}{I.~Elizabeth Kumar}.} \bibinfo{year}{2021}\natexlab{}.
\newblock \showarticletitle{Epistemic values in feature importance methods: Lessons from feminist epistemology}. In \bibinfo{booktitle}{\emph{FAccT '21: 2021 {ACM} Conference on Fairness, Accountability, and Transparency, Virtual Event / Toronto, Canada, March 3-10, 2021}}. \bibinfo{publisher}{{ACM}}, \bibinfo{pages}{817--826}.
\newblock


\bibitem[Huang and Marques-Silva(2024)]%
        {huang2024failings}
\bibfield{author}{\bibinfo{person}{Xuanxiang Huang} {and} \bibinfo{person}{Joao Marques-Silva}.} \bibinfo{year}{2024}\natexlab{}.
\newblock \showarticletitle{On the failings of Shapley values for explainability}.
\newblock \bibinfo{journal}{\emph{International Journal of Approximate Reasoning}} (\bibinfo{year}{2024}), \bibinfo{pages}{109112}.
\newblock


\bibitem[Ib{\'{a}}{\~{n}}ez and Olmeda(2022)]%
        {DBLP:journals/ais/IbanezO22}
\bibfield{author}{\bibinfo{person}{Javier~Camacho Ib{\'{a}}{\~{n}}ez} {and} \bibinfo{person}{M{\'{o}}nica~Villas Olmeda}.} \bibinfo{year}{2022}\natexlab{}.
\newblock \showarticletitle{Operationalising {AI} ethics: how are companies bridging the gap between practice and principles? An exploratory study}.
\newblock \bibinfo{journal}{\emph{{AI} Soc.}} \bibinfo{volume}{37}, \bibinfo{number}{4} (\bibinfo{year}{2022}), \bibinfo{pages}{1663--1687}.
\newblock


\bibitem[Ioannidis(2003)]%
        {DBLP:conf/vldb/Ioannidis03}
\bibfield{author}{\bibinfo{person}{Yannis~E. Ioannidis}.} \bibinfo{year}{2003}\natexlab{}.
\newblock \showarticletitle{The History of Histograms (abridged)}. In \bibinfo{booktitle}{\emph{Proceedings of 29th International Conference on Very Large Data Bases, {VLDB} 2003, Berlin, Germany, September 9-12, 2003}}. \bibinfo{publisher}{Morgan Kaufmann}, \bibinfo{pages}{19--30}.
\newblock


\bibitem[Jokanovic et~al\mbox{.}(2016)]%
        {jokanovic2016effect}
\bibfield{author}{\bibinfo{person}{Branka Jokanovic}, \bibinfo{person}{Moeness~G Amin}, {and} \bibinfo{person}{Fauzia Ahmad}.} \bibinfo{year}{2016}\natexlab{}.
\newblock \showarticletitle{Effect of data representations on deep learning in fall detection}. In \bibinfo{booktitle}{\emph{2016 IEEE Sensor Array and Multichannel Signal Processing Workshop (SAM)}}. IEEE, \bibinfo{pages}{1--5}.
\newblock


\bibitem[Kamiran and Calders(2011)]%
        {DBLP:journals/kais/KamiranC11}
\bibfield{author}{\bibinfo{person}{Faisal Kamiran} {and} \bibinfo{person}{Toon Calders}.} \bibinfo{year}{2011}\natexlab{}.
\newblock \showarticletitle{Data preprocessing techniques for classification without discrimination}.
\newblock \bibinfo{journal}{\emph{Knowl. Inf. Syst.}} \bibinfo{volume}{33}, \bibinfo{number}{1} (\bibinfo{year}{2011}), \bibinfo{pages}{1--33}.
\newblock


\bibitem[Laberge et~al\mbox{.}(2023)]%
        {DBLP:conf/iclr/LabergeA0MK23}
\bibfield{author}{\bibinfo{person}{Gabriel Laberge}, \bibinfo{person}{Ulrich A{\"{\i}}vodji}, \bibinfo{person}{Satoshi Hara}, \bibinfo{person}{Mario Marchand}, {and} \bibinfo{person}{Foutse Khomh}.} \bibinfo{year}{2023}\natexlab{}.
\newblock \showarticletitle{Fooling {SHAP} with Stealthily Biased Sampling}. In \bibinfo{booktitle}{\emph{The Eleventh International Conference on Learning Representations, {ICLR} 2023, Kigali, Rwanda, May 1-5, 2023}}. \bibinfo{publisher}{OpenReview.net}.
\newblock


\bibitem[Lundberg and Lee(2017)]%
        {DBLP:conf/nips/LundbergL17}
\bibfield{author}{\bibinfo{person}{Scott~M. Lundberg} {and} \bibinfo{person}{Su{-}In Lee}.} \bibinfo{year}{2017}\natexlab{}.
\newblock \showarticletitle{A Unified Approach to Interpreting Model Predictions}. In \bibinfo{booktitle}{\emph{Advances in Neural Information Processing Systems 30: Annual Conference on Neural Information Processing Systems 2017, December 4-9, 2017, Long Beach, CA, {USA}}}. \bibinfo{pages}{4765--4774}.
\newblock


\bibitem[Mockus(1974)]%
        {DBLP:conf/ifip7/Mockus74}
\bibfield{author}{\bibinfo{person}{Jonas Mockus}.} \bibinfo{year}{1974}\natexlab{}.
\newblock \showarticletitle{On Bayesian Methods for Seeking the Extremum}. In \bibinfo{booktitle}{\emph{Optimization Techniques, {IFIP} Technical Conference, Novosibirsk, USSR, July 1-7, 1974}} \emph{(\bibinfo{series}{Lecture Notes in Computer Science}, Vol.~\bibinfo{volume}{27})}. \bibinfo{pages}{400--404}.
\newblock


\bibitem[Ramachandranpillai et~al\mbox{.}(2023)]%
        {fairxai}
\bibfield{author}{\bibinfo{person}{Resmi Ramachandranpillai}, \bibinfo{person}{Ricardo Baeza-Yates}, {and} \bibinfo{person}{Fredrik Heintz}.} \bibinfo{year}{2023}\natexlab{}.
\newblock \bibinfo{title}{FairXAI - A Taxonomy and Framework for Fairness and Explainability Synergy in Machine Learning}.
\newblock
\newblock
\urldef\tempurl%
\url{https://doi.org/10.36227/techrxiv.24463945.v1}
\showDOI{\tempurl}


\bibitem[Ribeiro et~al\mbox{.}(2018)]%
        {DBLP:conf/aaai/Ribeiro0G18}
\bibfield{author}{\bibinfo{person}{Marco~T{\'{u}}lio Ribeiro}, \bibinfo{person}{Sameer Singh}, {and} \bibinfo{person}{Carlos Guestrin}.} \bibinfo{year}{2018}\natexlab{}.
\newblock \showarticletitle{Anchors: High-Precision Model-Agnostic Explanations}. In \bibinfo{booktitle}{\emph{Proceedings of the Thirty-Second {AAAI} Conference on Artificial Intelligence, (AAAI-18), the 30th innovative Applications of Artificial Intelligence (IAAI-18), and the 8th {AAAI} Symposium on Educational Advances in Artificial Intelligence (EAAI-18), New Orleans, Louisiana, USA, February 2-7, 2018}}. \bibinfo{publisher}{{AAAI} Press}, \bibinfo{pages}{1527--1535}.
\newblock


\bibitem[Robnik{-}Sikonja and Bohanec(2018)]%
        {DBLP:series/hci/Robnik-SikonjaB18}
\bibfield{author}{\bibinfo{person}{Marko Robnik{-}Sikonja} {and} \bibinfo{person}{Marko Bohanec}.} \bibinfo{year}{2018}\natexlab{}.
\newblock \showarticletitle{Perturbation-Based Explanations of Prediction Models}.
\newblock In \bibinfo{booktitle}{\emph{Human and Machine Learning - Visible, Explainable, Trustworthy and Transparent}}. \bibinfo{publisher}{Springer}, \bibinfo{pages}{159--175}.
\newblock


\bibitem[Saleiro et~al\mbox{.}(2018)]%
        {saleiro2018aequitas}
\bibfield{author}{\bibinfo{person}{Pedro Saleiro}, \bibinfo{person}{Benedict Kuester}, \bibinfo{person}{Loren Hinkson}, \bibinfo{person}{Jesse London}, \bibinfo{person}{Abby Stevens}, \bibinfo{person}{Ari Anisfeld}, \bibinfo{person}{Kit~T Rodolfa}, {and} \bibinfo{person}{Rayid Ghani}.} \bibinfo{year}{2018}\natexlab{}.
\newblock \showarticletitle{Aequitas: A bias and fairness audit toolkit}.
\newblock \bibinfo{journal}{\emph{arXiv preprint arXiv:1811.05577}} (\bibinfo{year}{2018}).
\newblock


\bibitem[Shapley et~al\mbox{.}(1953)]%
        {shapley1953}
\bibfield{author}{\bibinfo{person}{Lloyd~S Shapley} {et~al\mbox{.}}} \bibinfo{year}{1953}\natexlab{}.
\newblock \showarticletitle{A value for n-person games}.
\newblock  (\bibinfo{year}{1953}).
\newblock


\bibitem[Slack et~al\mbox{.}(2020)]%
        {DBLP:conf/aies/SlackHJSL20}
\bibfield{author}{\bibinfo{person}{Dylan Slack}, \bibinfo{person}{Sophie Hilgard}, \bibinfo{person}{Emily Jia}, \bibinfo{person}{Sameer Singh}, {and} \bibinfo{person}{Himabindu Lakkaraju}.} \bibinfo{year}{2020}\natexlab{}.
\newblock \showarticletitle{Fooling {LIME} and {SHAP:} Adversarial Attacks on Post hoc Explanation Methods}. In \bibinfo{booktitle}{\emph{{AIES} '20: {AAAI/ACM} Conference on AI, Ethics, and Society, New York, NY, USA, February 7-8, 2020}}. \bibinfo{publisher}{{ACM}}, \bibinfo{pages}{180--186}.
\newblock


\bibitem[Stoyanovich et~al\mbox{.}(2022)]%
        {DBLP:journals/cacm/StoyanovichAHJS22}
\bibfield{author}{\bibinfo{person}{Julia Stoyanovich}, \bibinfo{person}{Serge Abiteboul}, \bibinfo{person}{Bill Howe}, \bibinfo{person}{H.~V. Jagadish}, {and} \bibinfo{person}{Sebastian Schelter}.} \bibinfo{year}{2022}\natexlab{}.
\newblock \showarticletitle{Responsible data management}.
\newblock \bibinfo{journal}{\emph{Commun. {ACM}}} \bibinfo{volume}{65}, \bibinfo{number}{6} (\bibinfo{year}{2022}), \bibinfo{pages}{64--74}.
\newblock


\bibitem[Stoyanovich et~al\mbox{.}(2020)]%
        {NMI_2020}
\bibfield{author}{\bibinfo{person}{Julia Stoyanovich}, \bibinfo{person}{Jay~J. {Van Bavel}}, {and} \bibinfo{person}{Tessa~V. West}.} \bibinfo{year}{2020}\natexlab{}.
\newblock \showarticletitle{The Imperative of Interpretable Machines}.
\newblock \bibinfo{journal}{\emph{Nature Machine Intelligence}}  \bibinfo{volume}{2} (\bibinfo{year}{2020}), \bibinfo{pages}{197--199}.
\newblock
\urldef\tempurl%
\url{https://doi.org/10.1038/s42256-020-0171-8}
\showDOI{\tempurl}


\bibitem[Umm{-}e{-}Habiba et~al\mbox{.}(2025)]%
        {DBLP:journals/ese/UmmeHabibaHBFW25}
\bibfield{author}{\bibinfo{person}{Umm{-}e{-}Habiba}, \bibinfo{person}{Mohammad~Kasra Habib}, \bibinfo{person}{Justus Bogner}, \bibinfo{person}{Jonas Fritzsch}, {and} \bibinfo{person}{Stefan Wagner}.} \bibinfo{year}{2025}\natexlab{}.
\newblock \showarticletitle{How do {ML} practitioners perceive explainability? an interview study of practices and challenges}.
\newblock \bibinfo{journal}{\emph{Empir. Softw. Eng.}} \bibinfo{volume}{30}, \bibinfo{number}{1} (\bibinfo{year}{2025}), \bibinfo{pages}{18}.
\newblock


\bibitem[Vengroff(2024)]%
        {DBLP:conf/fat/Vengroff24}
\bibfield{author}{\bibinfo{person}{Darren~Erik Vengroff}.} \bibinfo{year}{2024}\natexlab{}.
\newblock \showarticletitle{Impact Charts: {A} Tool for Identifying Systematic Bias in Social Systems and Data}. In \bibinfo{booktitle}{\emph{The 2024 {ACM} Conference on Fairness, Accountability, and Transparency, FAccT 2024, Rio de Janeiro, Brazil, June 3-6, 2024}}. \bibinfo{publisher}{{ACM}}, \bibinfo{pages}{1187--1198}.
\newblock


\bibitem[Wang et~al\mbox{.}(2022)]%
        {wang2022towards}
\bibfield{author}{\bibinfo{person}{Angelina Wang}, \bibinfo{person}{Vikram~V Ramaswamy}, {and} \bibinfo{person}{Olga Russakovsky}.} \bibinfo{year}{2022}\natexlab{}.
\newblock \showarticletitle{Towards intersectionality in machine learning: Including more identities, handling underrepresentation, and performing evaluation}. In \bibinfo{booktitle}{\emph{Proceedings of the 2022 ACM Conference on Fairness, Accountability, and Transparency}}. \bibinfo{pages}{336--349}.
\newblock


\bibitem[Watson(2022)]%
        {DBLP:conf/fat/Watson22}
\bibfield{author}{\bibinfo{person}{David~S. Watson}.} \bibinfo{year}{2022}\natexlab{}.
\newblock \showarticletitle{Rational Shapley Values}. In \bibinfo{booktitle}{\emph{FAccT '22: 2022 {ACM} Conference on Fairness, Accountability, and Transparency, Seoul, Republic of Korea, June 21 - 24, 2022}}. \bibinfo{publisher}{{ACM}}, \bibinfo{pages}{1083--1094}.
\newblock


\bibitem[Wexler et~al\mbox{.}(2020)]%
        {DBLP:conf/fat/WexlerPRBZ20}
\bibfield{author}{\bibinfo{person}{James Wexler}, \bibinfo{person}{Mahima Pushkarna}, \bibinfo{person}{Sara Robinson}, \bibinfo{person}{Tolga Bolukbasi}, {and} \bibinfo{person}{Andrew Zaldivar}.} \bibinfo{year}{2020}\natexlab{}.
\newblock \showarticletitle{Probing {ML} models for fairness with the what-if tool and {SHAP:} hands-on tutorial}. In \bibinfo{booktitle}{\emph{FAT* '20: Conference on Fairness, Accountability, and Transparency, Barcelona, Spain, January 27-30, 2020}}. \bibinfo{publisher}{{ACM}}, \bibinfo{pages}{705}.
\newblock


\bibitem[Zafar et~al\mbox{.}(2019)]%
        {DBLP:journals/jmlr/ZafarVGG19}
\bibfield{author}{\bibinfo{person}{Muhammad~Bilal Zafar}, \bibinfo{person}{Isabel Valera}, \bibinfo{person}{Manuel Gomez{-}Rodriguez}, {and} \bibinfo{person}{Krishna~P. Gummadi}.} \bibinfo{year}{2019}\natexlab{}.
\newblock \showarticletitle{Fairness Constraints: {A} Flexible Approach for Fair Classification}.
\newblock \bibinfo{journal}{\emph{J. Mach. Learn. Res.}}  \bibinfo{volume}{20} (\bibinfo{year}{2019}), \bibinfo{pages}{75:1--75:42}.
\newblock


\bibitem[Zejnilovic et~al\mbox{.}(2021)]%
        {zejnilovic2021machine}
\bibfield{author}{\bibinfo{person}{Leid Zejnilovic}, \bibinfo{person}{Susana Lavado}, \bibinfo{person}{Carlos Soares}, \bibinfo{person}{{\'I}{\~n}igo Mart{\'\i}nez De Rituerto De~Troya}, \bibinfo{person}{Andrew Bell}, {and} \bibinfo{person}{Rayid Ghani}.} \bibinfo{year}{2021}\natexlab{}.
\newblock \showarticletitle{Machine learning informed decision-making with interpreted model’s outputs: A field intervention}. In \bibinfo{booktitle}{\emph{Academy of Management Proceedings}}, Vol.~\bibinfo{volume}{2021}. Academy of Management Briarcliff Manor, NY 10510, \bibinfo{pages}{15424}.
\newblock


\end{thebibliography}

\newpage
\appendix



%

\section{Local explanations}

\begin{figure*}[t]
    \begin{subfigure}[t]{0.48\textwidth}
        \includegraphics[width=\linewidth]{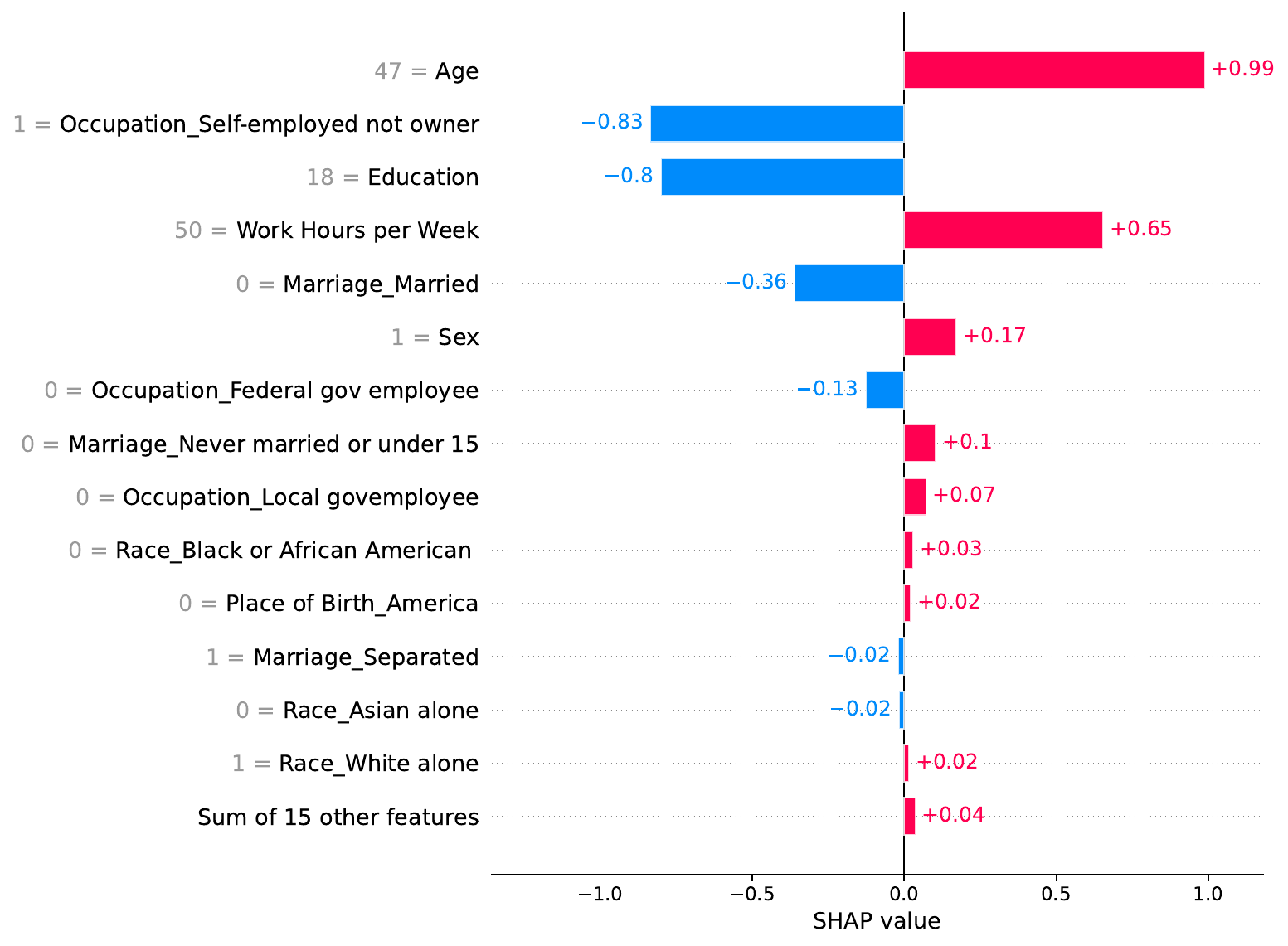}
        \caption{SHAP values before bucketization}
    \end{subfigure}
    \begin{subfigure}[t]{0.48\textwidth}
        \includegraphics[width=\linewidth]{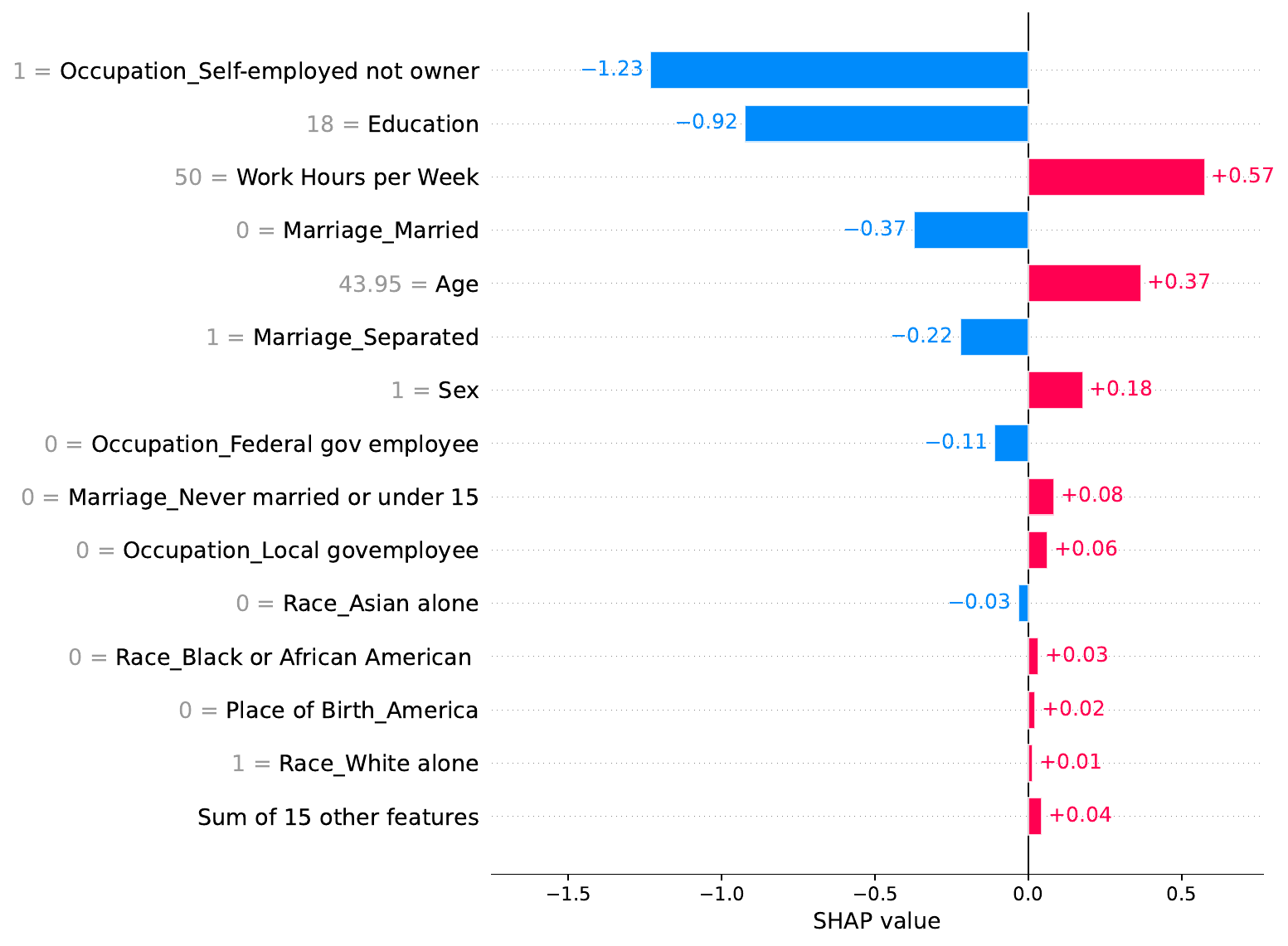}
        \caption{SHAP values after bucketizing age into 10 equally sized buckets}
    \end{subfigure} 
    \caption{SHAP values of features before (a) and after (b) bucketization for a fixed observation from ACS Public Coverage.}\label{fig:individualanalysis_full}
\end{figure*}

Continuing from Section~\ref{sec:introduction}, we perform individual analyses to understand how the SHAP ranking of age can decrease for a sample for the ACS Public Coverage dataset. Figure~\ref{fig:individualanalysis_full} shows how the age's SHAP ranking decreases for an individual. In (a), the age of the first individual is 22, which can be viewed as a small value. Once the age is bucketized into a range of ages, the value 22 is no longer special, and the age's SHAP value becomes negligible. 

\section{Further analyses on ACS Income}

\begin{figure*}[t]
    \begin{subfigure}[t]{0.33\textwidth}
        \includegraphics[width=\linewidth]{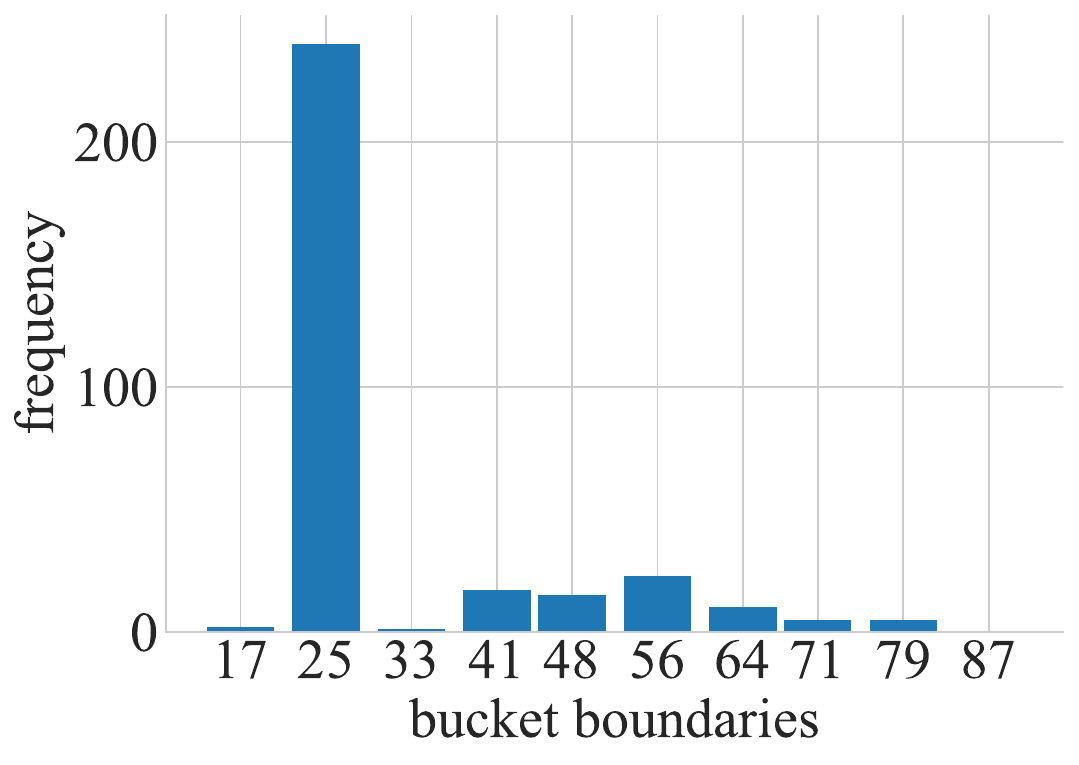}
        \caption{demotions in the rank of \emph{age}}
        \label{fig:sensitivity1b:d}
    \end{subfigure}
    \begin{subfigure}[t]{0.33\textwidth}
        \includegraphics[width=\linewidth]{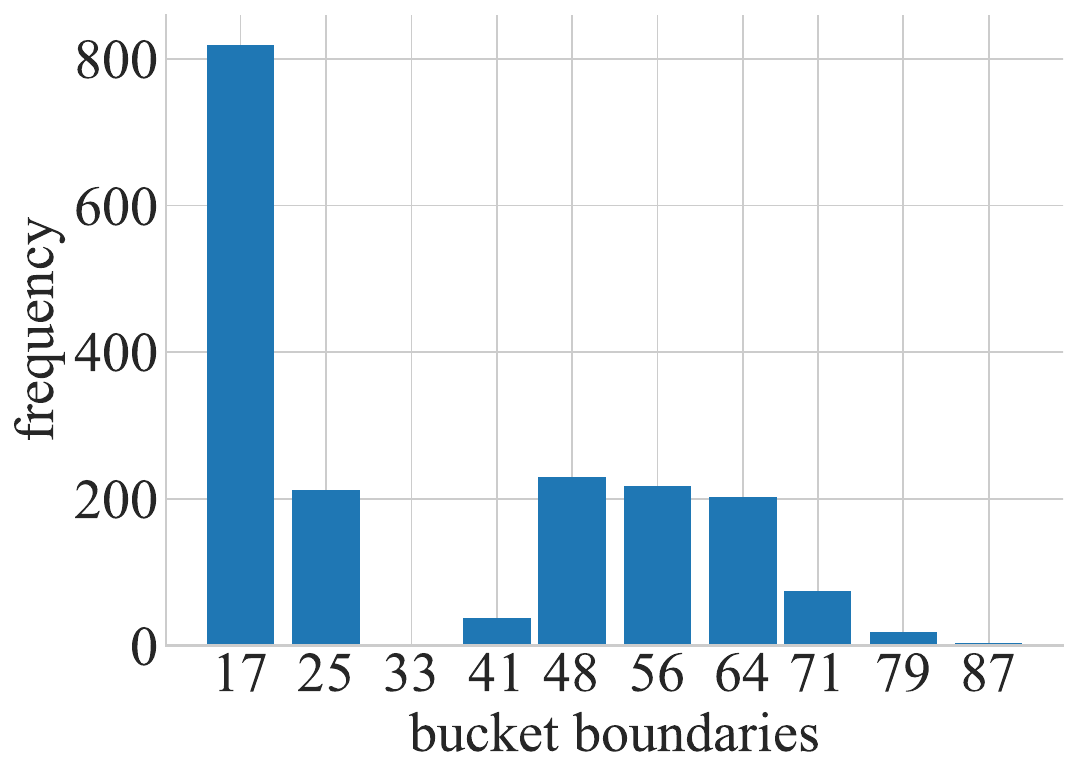}
        \caption{no change in the rank of \emph{age}}
        \label{fig:sensitivity1b:e}
    \end{subfigure} 

    \caption{
    Frequency plot for different buckets of the \emph{age} feature on ACS Income. Bucket boundaries are shown on the  $x$-axis.
    (a)  Number of observations where \emph{age} was the most important feature before bucketization, and where the relative importance of \emph{age} decreased (rank increased) after bucketization.  (b) Number of observations where \emph{age} was the most important feature both before and after bucketization. }
    
    \label{fig:sensitivity1b-appendix} 
\end{figure*}

Continuing from Section~\ref{sec:continuousfeaturesage}, we focus on individuals for whom \emph{age} is the highest-ranked (\ie most important) feature and show the number of observations for which the rank of the \emph{age} feature increased, decreased, or did not change in Figure~\ref{fig:sensitivity1b-appendix}. Only a few samples in the third bucket have age as the first rank.  

\begin{figure*}[t]
    \begin{subfigure}[t]{0.4\textwidth}
        \includegraphics[width=\linewidth]{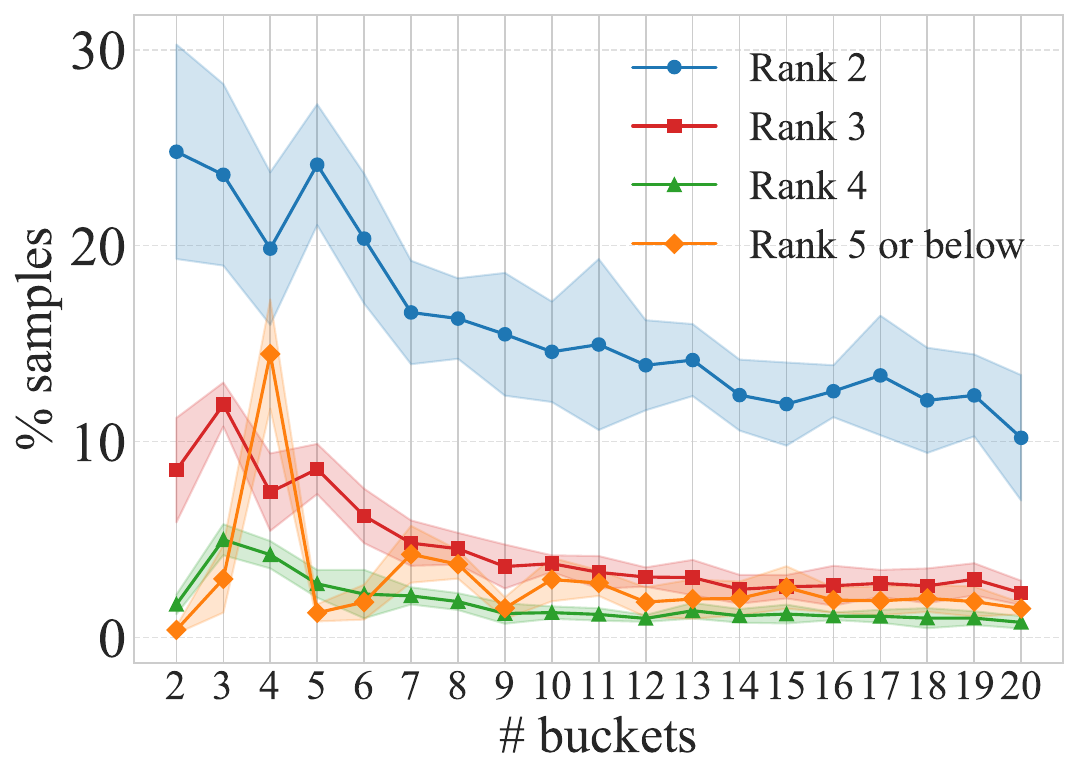}
        \caption{True positive}
        \label{fig:subgroups:a}
    \end{subfigure}
    \begin{subfigure}[t]{0.4\textwidth}
        \includegraphics[width=\linewidth]{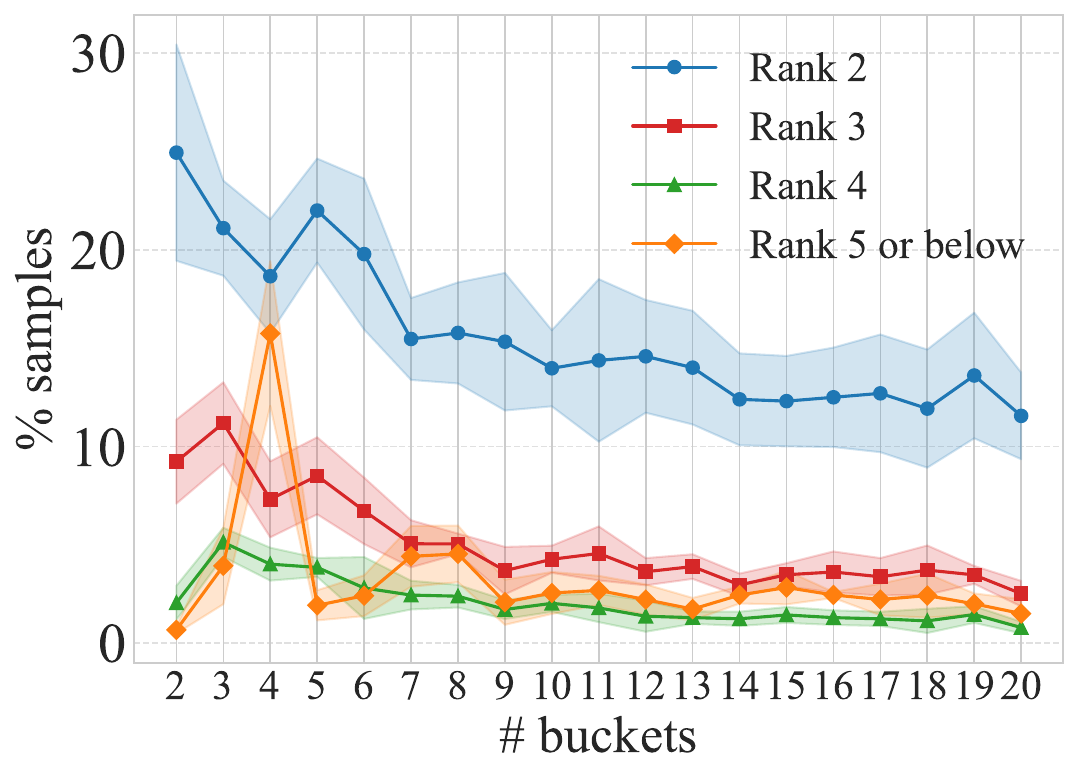}
        \caption{False positive}
        \label{fig:subgroups:b}
    \end{subfigure} \\
    \begin{subfigure}[t]{0.4\textwidth}
        \includegraphics[width=\linewidth]{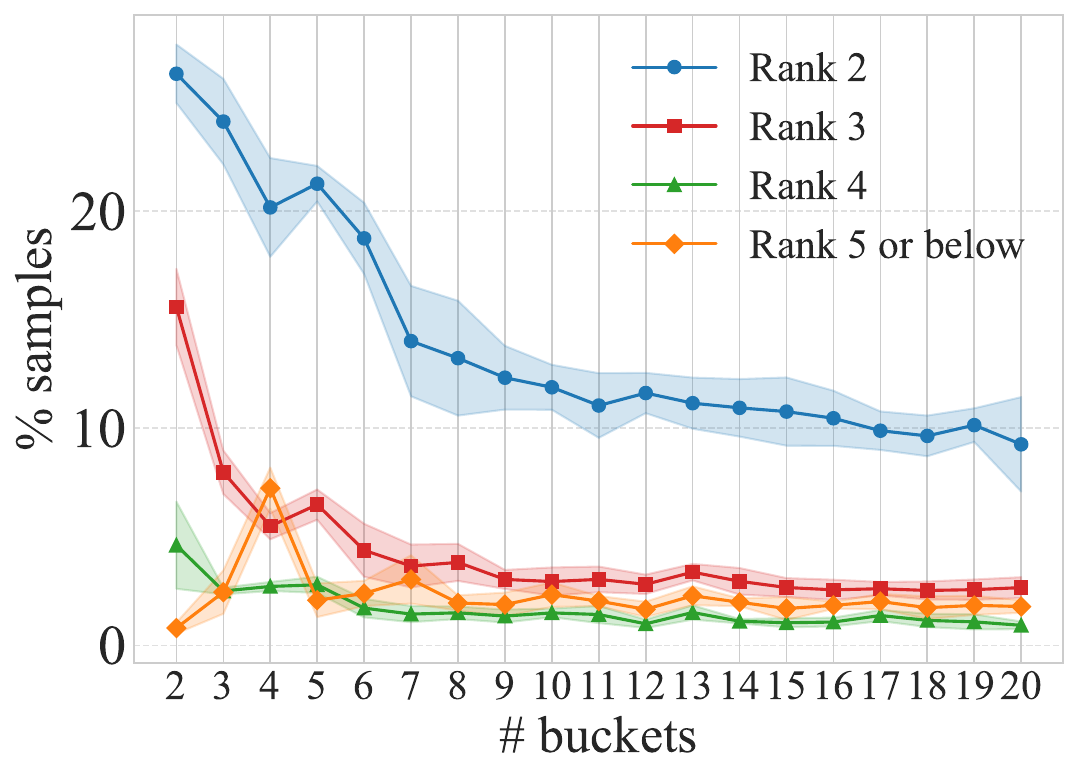}
        \caption{True negative}
        \label{fig:subgroups:c}
    \end{subfigure}
    \begin{subfigure}[t]{0.4\textwidth}
        \includegraphics[width=\linewidth]{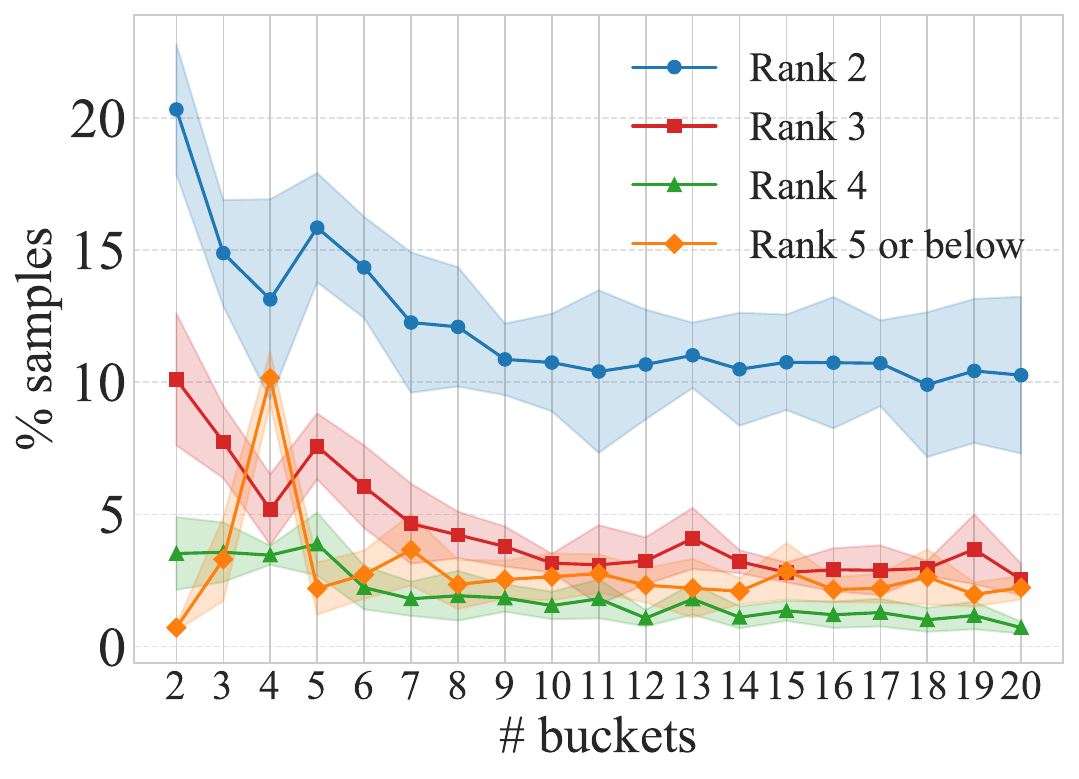}
        \caption{False negative}
        \label{fig:subgroups:d}
    \end{subfigure}
    \caption{SHAP ranking changes for the four outcomes where age was the highest-ranked feature as we change the number of buckets in the ACS Income dataset.}\label{fig:subgroups}
\end{figure*}

%

\section{SHAP Ranking Sensitivity}

Continuing from Section~\ref{sec:continuousfeaturesage}, we investigate how the SHAP ranking sensitivity changes based on the confusion matrix position of data points. Figure~\ref{fig:subgroups} shows that changes are not uniform across these groups.


\end{document}